\documentclass[journal, onecolumn]{IEEEtran}

\usepackage{cite}
\usepackage[flushleft]{threeparttable} 
\usepackage{booktabs}
\usepackage{graphicx}  
\usepackage[caption=false,font=footnotesize]{subfig}  
\usepackage{amsmath,amssymb,amsfonts}
\usepackage{mathtools}
\usepackage{algorithm}
\usepackage{algorithmic}
\usepackage{textcomp}
\usepackage{xcolor}
\usepackage{array} 
\usepackage{multirow} 
\usepackage{enumerate} 
\usepackage{soul} 

\usepackage{url}

\makeatletter
\IEEEtriggercmd{\reset@font\normalfont\scriptsize}
\makeatother
\IEEEtriggeratref{1}

\begin{document}
%

\title{A Federated Learning-based \\Industrial Health Prognostics for Heterogeneous Edge Devices using Matched Feature Extraction}

%
%

\author{Anushiya~Arunan,
        Yan~Qin,~\IEEEmembership{Member,~IEEE},
        Xiaoli~Li,~\IEEEmembership{Senior Member,~IEEE}, and Chau~Yuen,~\IEEEmembership{Fellow,~IEEE}
\thanks{This work was partly supported by the National Research Foundation, Singapore under its AI Singapore Programme (AISG Award No: AISG2-RP-2021-027). The work of Anushiya Arunan was supported by the A*STAR Graduate Scholarship. (Corresponding authors: Yan Qin and Chau Yuen)}
\thanks{A. Arunan is with the Engineering Product Development Pillar, Singapore University of Technology and Design, Singapore 487372, and also with the Institute for Infocomm Research and  Centre for Frontier AI Research, Agency for Science, Technology and Research (A*STAR), Singapore 138632 (e-mail: anushiya\_arunan@mymail.sutd.edu.sg).}
\thanks{Y. Qin is with the School of Automation, Chongqing University, Chongqing 400044, China, and also with the Engineering Product Development Pillar, Singapore University of Technology and Design, Singapore 487372 (e-mail: yan.qin@cqu.edu.cn).}
\thanks{X. Li is with both the Institute for Infocomm Research and Centre for Frontier AI Research, Agency for Science, Technology and Research (A*STAR), Singapore 138632 (e-mail: xlli@i2r.a-star.edu.sg).}
\thanks{C. Yuen is with the School of Electrical and Electronics Engineering,  Nanyang Technological University, Singapore 639798  (e-mail: chau.yuen@ntu.edu.sg).}}

%
%

\markboth{This paper has been accepted by IEEE Transactions on Automation Science and Engineering with DOI:10.1109/TASE.2023.3274648}
{Shell \MakeLowercase{\textit{\textit{et al.}}}: Bare Demo of IEEEtran.cls for IEEE Journals}
%



\maketitle
\begin{abstract}
Data-driven industrial health prognostics require rich training data to develop accurate and reliable predictive models. However, stringent data privacy laws and the abundance of edge industrial data necessitate decentralized data utilization. Thus, the industrial health prognostics field is well suited to significantly benefit from federated learning (FL), a decentralized and privacy-preserving learning technique. However, FL-based health prognostics tasks have hardly been investigated due to the complexities of meaningfully aggregating model parameters trained from heterogeneous data to form a high performing federated model. Specifically, data heterogeneity among edge devices, stemming from dissimilar degradation mechanisms and unequal dataset sizes, poses a critical statistical challenge for developing accurate federated models. We propose a pioneering FL-based health prognostic model with a feature similarity-matched parameter aggregation algorithm to discriminatingly learn from heterogeneous edge data. The algorithm searches across the heterogeneous locally trained models and matches neurons with probabilistically similar feature extraction functions first, before selectively averaging them to form the federated model parameters. As the algorithm only averages similar neurons, as opposed to conventional naive averaging of coordinate-wise neurons, the distinct feature extractors of local models are carried over with less dilution to the resultant federated model. Using both cyclic degradation data of Li-ion batteries and non-cyclic data of turbofan engines, we demonstrate that the proposed method yields accuracy improvements as high as 44.5\% and 39.3\% for state-of-health estimation and remaining useful life estimation, respectively.
\end{abstract}



%
\IEEEpeerreviewmaketitle

\section{Introduction}
\IEEEPARstart{E}{dge} devices with powerful sensing and computing capabilities have grown from being a novelty to a utility, permeating all aspects of the modern industry \cite{munirathinam2020industry}. By bringing the benefits of artificial intelligence and the Internet of Things closer to the end user for real-time analytics and decision making, edge devices play a critical role in Industry 4.0. Thus, proactive health management of these assets, through condition monitoring and timely predictive maintenance, is vital to prevent costly failures and ensure reliable operations. Life cycle costs of operating edge assets are also reduced with condition-based maintenance as a cost-effective balance between unplanned reactive maintenance and excessive preventive maintenance can be achieved \cite{mobley2002introduction}.

Consequently, developing industrial health prognostic models to guide condition-based maintenance has attracted significant scientific and industry interest, with a growing emphasis on data-driven models that can leverage the real-time data generated by edge devices \cite{lei2018machinery, guo2020process, xu2023hybrid, pang2021age}. Amongst data-driven methods, deep learning models have gained notable prominence due to their ability build accurate models from raw input data without needing extensive domain knowledge \cite{yao2022survey}\cite{dong2021modeling}. For instance, Chen \textit{et al.} \cite{chen2020machine} developed an attention-enhanced long short-term memory (LSTM) model trained on sensor data to estimate the remaining useful life (RUL) of turbofan engines. Zhang \textit{et al.} \cite{zhang2020time} utilized discharging voltage and current data generated from a convolutional recurrent generative adversarial network to build a state-of-health (SoH) estimation model for Li-ion batteries. Zhu \textit{et al.} \cite{zhu2018estimation} proposed a multiscale convolutional neural network (CNN) for the RUL estimation of rolling bearings using vibration signals as model inputs.

However, an underlying assumption in existing studies is that the decentralized data from multiple devices can be centrally pooled for model training. In real industries, aggregating data is often unviable given stringent data privacy laws, such as the General Data Protection Regulation \cite{nguyen2021federated}, which limits the use and dissemination of sensitive user data. For example, battery health data from electric vehicles (EV) is beneficial to an EV manufacturer for the optimal selection of long-lasting battery types. However, collating battery health data generated from EV drivers is challenging as private user information such as driving patterns may be gleaned from just battery consumption patterns \cite{lee2019learning}. As a result, the wealth of data collectively held across edge devices cannot be harnessed in practice to train robust health prognosis models, covering a diverse range of practical operations and degradation patterns. Since deep learning models' performance typically increases with data quantity and quality \cite{adadi2021survey}, a lack of access to rich edge data may cap productivity gains from intelligent condition monitoring in Industry 4.0 \cite{mobley2002introduction} \cite{li2020review}.

To meet the demand for privacy-preserving machine learning, several techniques have emerged, and they can be broadly classified into data masking-based or architectural approaches \cite{xu2021privacy}. Data masking is typically utilized when raw data is transmitted to third parties or a central server for model training \cite{yang2019federated}. A few notable data masking techniques are differential privacy \cite{dwork2011firm} \cite{dwork2014algorithmic}, which apply well-designed noise perturbations to obfuscate sensitive individual data attributes, and homomorphic encryption \cite{rivest1978data}\cite{li2020privacy}, which enable computations to be performed directly on encrypted data without having to decrypt it to the original form first.

In contrast, architectural approaches to privacy-preservation focus on redesigning centralized machine learning paradigms \cite{xu2021privacy}. Among architectural methods, federated learning (FL)\cite{mcmahan2017communication} has garnered considerable research attention \cite{46432,li2020federated, wang2020federated, wang2020tackling} as it facilitates both privacy preservation and decentralized data utilization simultaneously. This makes FL highly relevant for the problem of learning from edge data under privacy constraints, and is thus explored further in this paper. In FL, local data owners, also termed clients, collaboratively train a federated model without disclosing their data. Clients train their own models at the edge using their available data, and only model parameters (i.e., weights and biases) of the client models are communicated to the central server to build the federated model. The data remains with each client, thereby ensuring data privacy \cite{mcmahan2017communication}. Presently, the most benchmarked FL algorithm is Federated Averaging (FedAvg)  \cite{mcmahan2017communication}, where local model parameters are aggregated by computing a coordinate-wise weighted average of the client models' parameters.

In existing works employing FL for machine condition monitoring, Zhang and Li \cite{zhang2022data} utilized FedAvg and transfer learning for fault classification of rolling bearings operating under variable conditions. Liu \textit{et al.} \cite{liu2020deep}  similarly used FedAvg to construct a federated LSTM-based anomaly detector for engine operations. Going beyond fault classification, Rosero \textit{et al.} \cite{rosero2020remaining} applied FL for regression problems, where they compared the performance of federated models based on FedAvg and its variant, FedProx \cite{li2020federated} for RUL estimation of turbofan engines. In the aforementioned studies, though the underlying model architectures for the various condition monitoring tasks differed, FedAvg—with its coordinate-wise model parameter averaging scheme—was commonly used for developing the federated model.

However, a shortcoming of FedAvg is that it implicitly assumes coordinate-wise neurons of different client models are learning similar features and hence, they can be averaged. In reality, even with homogeneous datasets across clients, there is no guarantee that the first neuron of a client model is learning the same feature extractors as the first neuron of another client model, and so on \cite{wang2020federated}. Amidst data heterogeneity, the problem of averaging dissimilar, coordinate-wise neurons is exacerbated, with Wang \textit{et al.} \cite{wang2020tackling} demonstrating that FedAvg converges to sub-optimal solutions when there are mismatched objective functions of different clients undergoing heterogeneous learning steps.

In a privacy-concerned era, intelligent health prognostics models can benefit significantly from FL because of the exposure to a wider range of operational and machine failure data than any single client’s dataset. However, employing FL techniques on industrial data is challenging due to inherent attributes of industrial data. Industrial data, typically sensor data, is noisy and temporal in nature. Consequently, neural networks such as LSTM, which effectively capture important long-term dependencies in long sequence time series \cite{hochreiter1997long}, tend to have higher prediction accuracies in health prognostics than  shallow architectures such as support vector machines \cite{zheng2017long, zhang2018long,zhao2016machine}, and even other standard deep networks such as CNN \cite{zheng2017long}\cite{choi2019machine} and gated recurrent units (GRU) \cite{park2020lstm, zhou2019remaining, ungurean2020online}. Thus, suitable FL techniques beyond naive parameter averaging are needed to ensure that the weights and biases of hidden states are aggregated meaningfully for the integrity of the overall federated model. Moreover, FL techniques also need to account for heterogeneous data across clients, even for devices of the same type. Data heterogeneity can arise among clients as machines experience dynamic and dissimilar degradation processes in practical conditions, and individual clients may have unequal training data (unbalanced data) from differing machine usage or data collection patterns.

Overall, the crucial research gaps that need to be tackled for reliable industrial health prognostics in a privacy-constrained era are summarized below:
\vspace{-0.2em}
\begin{itemize}
  \item To realize the theorized benefits of
intelligent health prognostics, in-depth FL studies are urgently needed for complex, time series regression-based condition monitoring and health prognosis problems. Otherwise, rich edge data remain untapped, and productivity gains from intelligent condition monitoring are capped in Industry
4.0 \cite{mobley2002introduction}\cite{li2020review}.
\vspace{0.2em}
  \item Data heterogeneity poses a critical statistical challenge for reliable FL as parameter averaging methods that overlook local heterogeneity dampens the federated model’s feature extraction capability and consequently, its predictive performance. Thus, there is a fundamental need for investigating new parameter aggregation algorithms that are suited to both the inherent heterogeneity of degradation data and the recurrent neural networks  used for industrial health prognosis.
\end{itemize}

In view of the challenges posed by client heterogeneity, we postulate that a more suitable parameter aggregation method for industrial health prognostics is one that explicitly considers each client model neuron's feature extraction (i.e., features learnt from input data) and selectively aggregates the neuron parameters based on the similarity of extracted features. Thus, drawing inspiration from the matched averaging (FedMA) algorithm \cite{wang2020federated} originally introduced for computer vision and language modeling, we develop a pioneering federated model for time series regression-based industrial health prognostics, which efficiently matches similar feature extractions from heterogeneous degradation data for a more discriminating parameter averaging. Specifically, the similarity between client model neurons is determined by their posterior probability distributions, and optimal matching of similar neurons is achieved by solving a neuron assignment cost matrix, formulated as a linear sum assignment problem. Since the FedMA algorithm only averages similar neurons, as opposed to FedAvg’s coordinate-wise averaging of potentially dissimilar neurons, the distinct feature extractors of client models are carried over to the federated model with less dilution, thereby boosting its predictive capabilities.

As the seminal FedMA algorithm is designed for complex models with domain-specific components, such as two-dimensional convolution layers or word embedding layers, its applicability to industrial health prognostic models is not straightforwardly clear. Thus, to better suit the lightweight prognostic models used in edge devices, we introduce a customized FedMA algorithm that directly targets the core components of the prognostic model, e.g., temporal hidden states. Furthermore, as we implement the matched averaging of neurons modularly, one model layer at a time, the proposed method copes well with deep architectures, such as LSTM networks that are popularly used for industrial health prognosis. On the whole, our proposed FedMA-based health prognostics model is pivotal for demonstrating how the matched averaging algorithm, previously unconsidered for industrial health prognostics, can be formulated within an
end-to-end health prognostics framework, consisting of both offline federated model training and online prognostics on unseen test data.

Overall, the main contributions of this paper are:
\begin{enumerate}[i)]
  \item We present a systematic empirical investigation of FL algorithms for time series regression-based industrial health prognosis that handles both cyclic and non-cyclic degradation data in a unified framework.
  \item We fundamentally rethink the default coordinate-wise parameter averaging for FL-based health prognostics, and propose a feature similarity-matched averaging scheme for a more discriminating parameter aggregation under practical heterogeneous conditions.
  \item We validate the effectiveness of our federated prognostic model on two different industrial datasets, achieving robust accuracy improvements as high as 44.5\% and 39.3\% for SoH and RUL estimation, respectively. Thus, our work is pivotal for spurring research into effective parameter aggregation methods for FL-based industrial health prognostic models.
\end{enumerate}

The remainder of the paper is organized as follows. Section II describes some preliminary information for a better understanding of industrial health prognosis and federated learning. Section III  formulates the proposed federated prognostic model with the feature similarity-matched  parameter averaging scheme. Section IV discusses the dataset, experimental formulation, and results. Section V concludes and offers future research directions.

\begin{figure}[!t]
\centering
\includegraphics[width=0.6\linewidth]{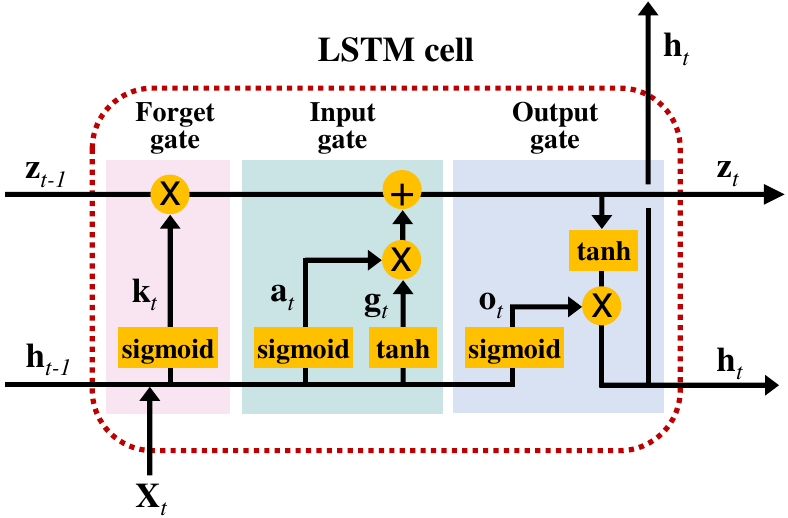}
\vspace{-1em}
\caption{Inner structure of a LSTM cell and the information flow within it.}
\vspace{-1.2 em}
\label{LSTMnetwork}
\end{figure}

\section{Preliminary}
To aid the understanding of the advanced FL-based prognostic models proposed in the paper, this section discusses some key background knowledge on the quintessential health prognostic tasks encountered for cyclic and non-cyclic degradation patterns, the basic LSTM architecture utilized for health modeling, and the standard federated model development process.

\subsection{Health Prognosis for Cyclic and Non-cyclic Degradation}
Degradation patterns can be broadly classified into cyclic and non-cyclic. Cyclic degradation refers to non-monotonic and repetitive degradation behavior from cyclical stresses. For instance, Li-ion batteries experience cyclic capacity degradation due to electrochemical losses from repeated charging and discharging operations \cite{arora1998capacity}. Thus, a quintessential health prognosis task for batteries is SoH estimation, which predicts a battery's practically achievable capacity for future operating cycles till its end-of-life.

In contrast, non-cyclic degradation patterns are monotonous and non-repetitive. A well-studied example is the piecewise degradation modeling of turbofan engines \cite{chen2020machine}, where degradation is negligible during initial operations, but starts to occur continuously after some time in operation. For these industrial assets, RUL estimation is a health prognosis task of significant interest, where RUL is defined as the remaining length of operational time until an asset’s complete failure.

\subsection{LSTM Neural Network for SoH and RUL Estimation}
As an architecture well-suited for long sequence prediction, LSTM is commonly used in health prognostic models of critical assets, such as engines \cite{zheng2017long}\cite{zhang2018long}, bearings \cite{dong2023deep, shen2022remaining, guo2021remaining}, and batteries \cite{ qin2021transfer, chemali2017long, 10018489}. The effectiveness of LSTM is due to its ability to keep track of important sequential dependencies in input data, even across long time periods \cite{hochreiter1997long}. Compared to the structurally simpler  GRU network \cite{cho-etal-2014-properties}, LSTM also performs better on complex sequences and learning tasks due to its enhanced information retention abilities \cite{cahuantzi2021comparison, irie2016lstm,gruber2020gru}. Thus, LSTM is used as the backbone for the proposed SoH and RUL estimation models in this paper.

A basic LSTM cell is an information holding unit with three gates (forget, input, and output) that allow important information to be propagated across time and irrelevant information to be forgotten. As shown in Fig. \ref{LSTMnetwork}, at each time $t$, the current input sequence $\mathbf{X}_t$ and the previous hidden state $\mathbf{h}_{t-1}$ are fed into gates with sigmoid or tanh functions to yield intermediate outputs, $\mathbf{k}_t$, $\mathbf{a}_t$, $\mathbf{g}_t$, and $\mathbf{o}_t$. As the elements of these outputs range from 0 to 1 (for sigmoid activation) and -1 to 1 (for tanh activation), they are essentially scaled values indicating the relative importance of the information in $\mathbf{X}_t$ and $\mathbf{h}_{t-1}$. These  values are then used to scale the relative contribution of the long-term information that is held in the previous cell state $\mathbf{z}_{t-1}$ to determine the new cell state $\mathbf{z}_{t}$ and hidden state $\mathbf{h}_{t}$, as follows:
\begin{equation}
\begin{aligned}
& \mathbf{z}_t=\mathbf{k}_t \times \mathbf{z}_{t-1}+\mathbf{a}_t \times \mathbf{g}_t \\
& \mathbf{h}_t=\mathbf{o}_t \times \tanh \left(\mathbf{z}_t\right)
\end{aligned}
\end{equation}
\noindent where $\times$ refers to element-wise multiplication.

The final predicted output $\widehat{y}_t$ (e.g., RUL values) is obtained from $ \widehat{y}_t = \sigma\left(\mathbf{h}_t\right)$, where $\sigma(.)$ refers to fully connected regression layers with appropriate activation functions.

\begin{figure}[!t]
\centering
\includegraphics[width=0.6\linewidth]{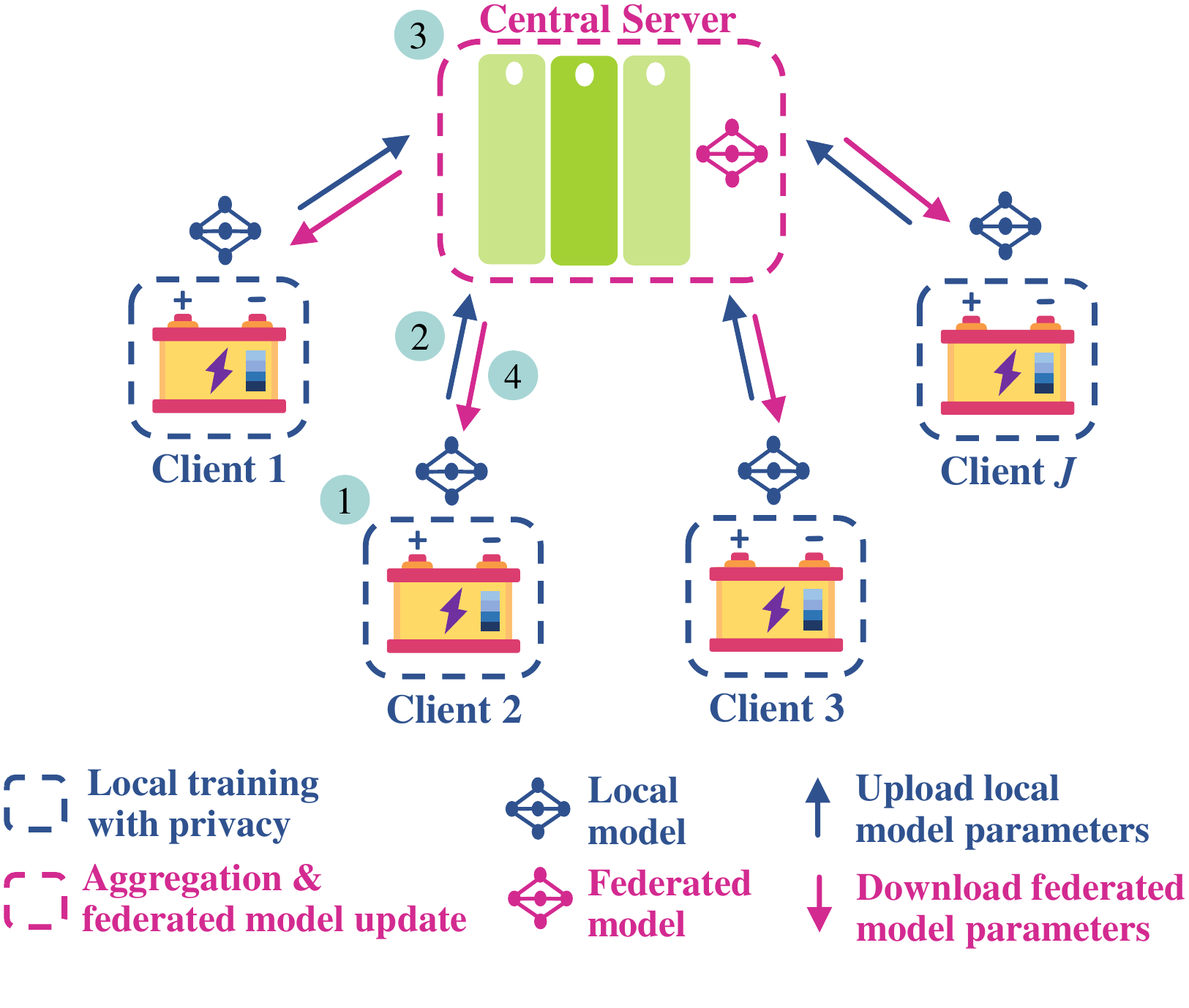}
\vspace{-1.7em}
\caption{Key steps of federated learning: \raisebox{.5pt}{\textcircled{\raisebox{-.9pt}{1}}} local model training, \raisebox{.5pt}{\textcircled{\raisebox{-.9pt} {2}}} local model parameter upload, \raisebox{.5pt}{\textcircled{\raisebox{-.9pt} {3}}} parameter aggregation to form federated model, and \raisebox{.5pt}{\textcircled{\raisebox{-.9pt} {4}}} adoption of federated model by clients.}
\vspace{-1.2 em}
\label{FLschematic}
\end{figure}

\begin{figure*}[!t]
\centering
\includegraphics[width=0.99\textwidth]{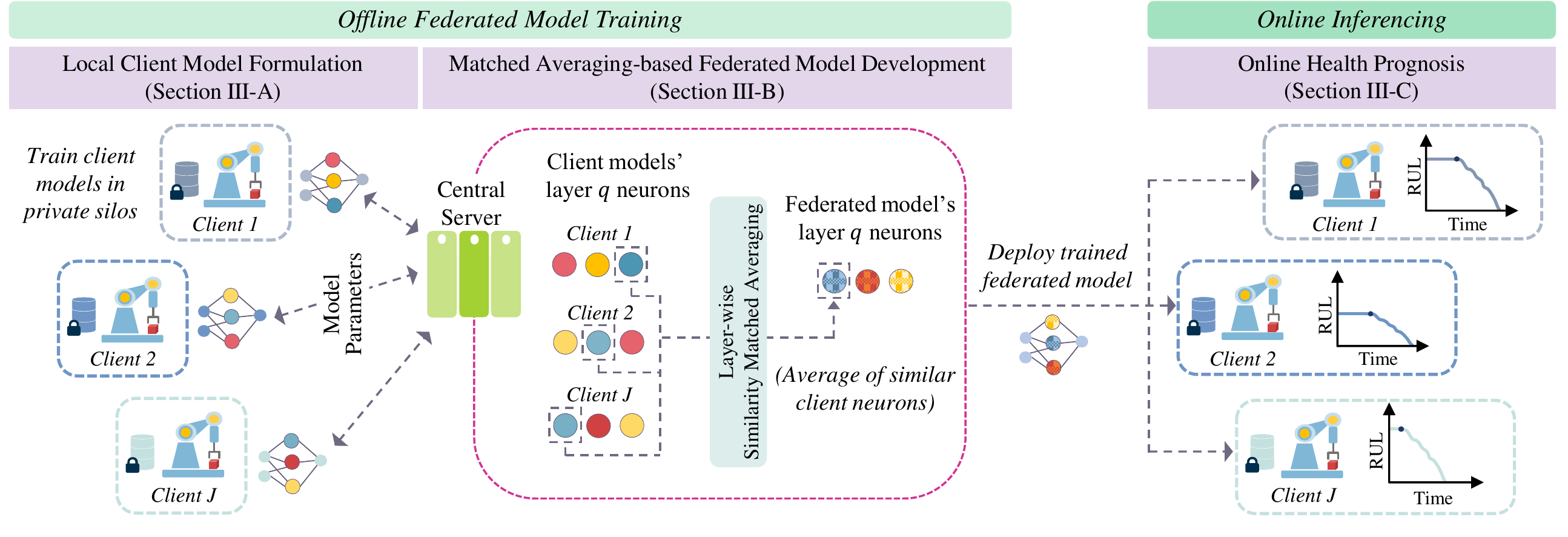}
\vspace{-1.7em}
\caption{Overview of proposed FedMA-based health prognosis framework.}
\vspace{-1.2 em}
\label{processflow}
\end{figure*}

\subsection{Federated Learning}
The goal of FL is to build a shared federated model that is attuned to local client-specific features and yet, also general enough to handle the diverse practical conditions not seen by a local training dataset. Since our focus is \textit{horizontal FL} \cite{nguyen2021federated}, where client data have the same feature and target variables but differ in the sample space, the same model architecture can be adopted for all client models and correspondingly, the federated model. The standard FL process is summarized as a high-level schematic in Fig. \ref{FLschematic}, and discussed below, using the seminal FedAvg algorithm as an example.

\begin{enumerate}[1)]
  \item Local clients independently train their individual local models for $E$ epochs with their private data.
  \item The trained local model parameters are communicated to a central server, and aggregated to form the parameters of a new federated model or update an existing one.
  \item For FedAvg, the local model parameters are averaged by computing a coordinate-wise mean, weighted by the proportion of the client dataset size:
\vspace{-0.5 em}
\begin{equation}
\mathbf{W}^{r} = \sum_{j=1}^{J}p_{j} \mathbf{W}_{j}^{r}
\vspace{-0.4 em}
\end{equation}
\noindent where $\mathbf{W}_j^r$ is client $j$'s model parameters at communication round $r$, $p_j$ is the fraction of total data samples belonging to client $j$, and $\mathbf{W}^{r}$ is the  federated model parameters.
  \item The federated model parameters $\mathbf{W}^{r}$ are broadcasted back to the local clients, and their models are initialized with the federated model parameters.
\end{enumerate}

\section{FL-based Industrial Health Prognostics using Matched Feature Extraction}
The proposed FedMA-based health prognostics framework consists of two stages—offline federated model training with idle local clients and online inferencing using the trained model. Individual clients kickoff the offline modeling process by training their respective initial client models at the edge, using only their private data. To then build the federated model, a layer-wise matched averaging algorithm is applied to the initial client models, where the hidden neurons of different clients are matched and aligned based on similar feature extraction traits first before their weights and biases are averaged. These averaged  parameters become the federated model’s parameters, which are subsequently distributed back to the clients for further rounds of collaborative federated model training. During online inferencing, clients employ the trained federated model for health prognosis tasks on incoming private test data. The steps of the proposed method are summarized in Fig. \ref{processflow}.

\subsection{Local Client Model Formulation}
As the local model architecture serves as the basis for the federated model in \textit{horizontal FL}, we begin with the local model formulation for some quintessential health prognosis tasks, e.g., SoH estimation or RUL estimation. Since the model architecture for each type of prognostic task is the same across all clients (only parameter values differ), a general $Q$-layer model is discussed from a single client $j$’s perspective. Here, $Q$ is the cardinality of the set of layer indexes for a general LSTM-based architecture depicted in Table \ref{Table_LSTM architecture}.

\begin{table}
\newcolumntype{P}[1]{>{\centering\arraybackslash}p{#1}}
\renewcommand{\arraystretch}{1.3}
\scriptsize
\centering
\caption{Parameters of a Q-layer LSTM-based model architecture}
\label{Table_LSTM architecture}
\begin{threeparttable}
\begin{tabular}{P{1.2cm}P{2.7cm}P{2cm}}
\hline
Layer index & Parameter, $W_{j,q}$ & Shape\tnote{a} \\
\hline
\multirow{4}{*}{1} & lstm.weight.input-hidden & ($U*L$, $D_{in}$)\\
 & lstm.weight.hidden-hidden & ($U*L$, $L$)\\
 & lstm.bias.input-hidden  &  ($U*L$)\\
 & lstm.bias.hidden-hidden &  ($U*L$)\\
$q$ & \tiny $\vdots$ & \tiny $\vdots$ \\
\multirow{2}{*}{$Q$} & regressor.weight & ($D_{out}$, $L$)\\
 & regressor.bias & ($D_{out}$)\\
\hline
\end{tabular}
\begin{tablenotes}
\item[a] $U$, $D_{in}$, $L$, and $D_{out}$ are, respectively, the number of intermediate LSTM cell weights, input features, hidden neurons, and output samples.
\end{tablenotes}
\end{threeparttable}
\vspace{-1.2 em}
\end{table}

\subsubsection{SoH Estimation with Cyclic Degradation}
Assuming $M$ process variables (e.g., voltage, temperature of discharging battery) are measured at discrete time indexes $t \in\left\{1, \ldots, T_s\right\}$ for each operating cycle $s$  in the cyclic data of client $j$, a feature data matrix $\mathbf{X}_s \in \mathbb{R}^{M \times T_s}$ can be formed, where $T_s$ is the varying cycle length of different cycles. To develop an LSTM-based SoH estimation model, the feature data matrix has to be transformed into inputs that are suited for the LSTM architecture. First, the feature data are standardized to have zero mean and unit variance with respect to each process variable so that extreme values do not skew model training and convergence. Next, as the LSTM architecture requires fixed-length input sequences, the standardized feature data are segmented into smaller sequences determined by the shortest cycle to become $\overline{\mathbf{X}}_s \in \mathbb{R}^{M \times T}$, where scalar $T=\min\left\{T_s\right\}_{s=1}^S$. The health indicator label (e.g., battery capacity) at the end of each cycle is denoted as $Y_s$.

The pairwise sequences $\left(\overline{\mathbf{X}}_s, Y_s\right)$ are then consecutively fed into an LSTM network to train client $j$'s SoH estimation model, $\widehat{Y}_s = \varphi\left(\overline{\mathbf{X}}_s, \mathbf{W}_j\right)$, where $\varphi$ is a non-linear function and $\mathbf{W}_j=\left\{\mathbf{W}_{j,q}\right\}_{q=1}^Q$, the set of model parameters for the $Q$ layers. The optimal $\mathbf{W}_j$ is obtained by minimizing the estimation error:
\begin{equation}
\min_{\mathbf{W}_j} \sum_{s=1}^S\left\|Y_s -\varphi\left(\overline{\mathbf{X}}_s, \mathbf{W}_j\right)\right\|_2^2
\end{equation}
\subsubsection{RUL Estimation with Non-cyclic Degradation}
For non-cyclic degradation, the progression of degradation occurs continuously across time, and there are no clear repetitive degradation patterns across operational cycles. Thus, the feature data matrix is represented without any reference to cycles as $\mathbf{X} \in \mathbb{R}^{M \times \mathcal{T}}$ , where $\mathcal{T}$ is length of the time series measurements of the $M$ process variables. For run-to-failure data, $\mathcal{T}$ is equivalent to the lifespan of the device. The feature data pre-processing steps of standardization (zero mean and unit variance) and data segmentation into smaller, fixed-length sequences are similarly performed.

For the data segmentation, a sliding window approach is utilized since non-cyclic degradation data need not be segmented by cycles\cite{9762891}. A sliding window of length $\lambda$ is shifted through the entire time series with time step size $\Delta \tau=1$ to generate smaller sequences $\widetilde{\mathbf{X}}[\tau: \tau+\lambda]$, starting from time index $\tau$ and of length $\lambda$. The corresponding health indicator label (i.e., RUL) at the end of the feature data sequence is given as $Y[\tau+\lambda]$. The pairwise segmented sequences, still denoted as $\left(\widetilde{\mathbf{X}}, Y\right)$ for brevity, are consecutively fed into the LSTM network to train client $j$’s $Q$-layer RUL estimation model, $\widehat{Y} = \varphi\left(\widetilde{\mathbf{X}}, \mathbf{W}_j\right)$, where $\mathbf{W}_j$ is once again the set of model parameters for the $Q$ layers. The optimal $\mathbf{W}_j$ is obtained by minimizing the estimation error, similar to Eq. 3.

\subsection{Matched Averaging-based Federated Model Development}
To build a potentially higher performing health prognostic model than any single local model trained on limited private data, a $Q$-layer federated model that learns from the optimal parameters $\left\{\mathbf{W}_j\right\}_{j=1}^J$  of $J$ clients is developed in this section. To aggregate the local model parameters meaningfully in the presence of client heterogeneity, neurons of different client models are considered layer-by-layer and matched based on their feature extraction similarity first before their parameters are averaged to form the federated model’s parameters. The mathematical formulation for the similarity-based neuron matching and the layer-wise matched averaging process for the federated model development are discussed below.

\subsubsection{Similarity-based Neuron Matching}
\begin{figure}[!t]
\vspace{-1.2 em}
\centering
\subfloat[]{\includegraphics[width=0.3\linewidth]{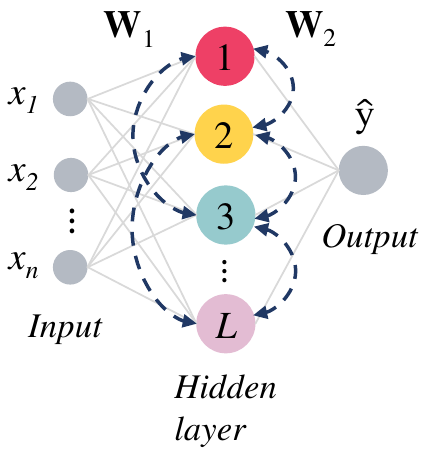}%
\label{permutation invariance}}
\hfil
\subfloat[]{\includegraphics[width=0.5\linewidth]{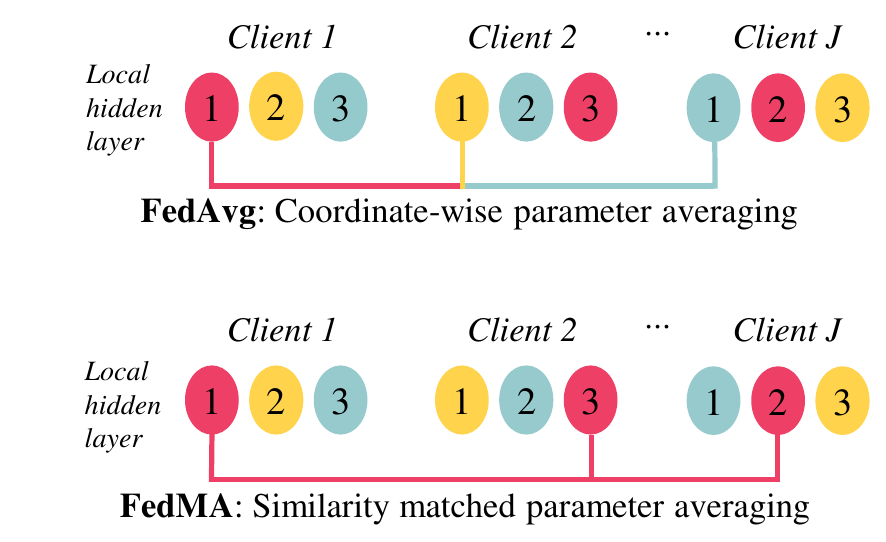}%
\label{fedavg_vs_ma}}
\caption{(a) A basic fully connected neural network has $L$! equivalent parametrizations due to the permutation invariance of parameter ordering. (b) Visual comparison of FedAvg vs. FedMA parameter averaging scheme, where similar neurons of different clients are represented by the same color.}
\vspace{-1.6em}
\label{fedma_buildup}
\end{figure}

The need for matching neurons based on similar feature extraction traits arises due to permutation invariance of neural network parameters. For a given network, there exists several equivalent variants that differ in the ordering of model parameters but produce the same output value \cite{wang2020federated}. For example, in the basic fully connected client model illustrated in Fig. \ref{fedma_buildup}\subref{permutation invariance}, $\hat{y}=\sum_{i=1}^L \mathbf{W}_{j,2,i \cdot}  \sigma\left(\left\langle x, \mathbf{W}_{j,1, \cdot i}\right\rangle\right)$, where $\cdot i$ and $i \cdot$ are the $i$-th column and $i$-th row of weights $\mathbf{W}_{j,1}$ and $\mathbf{W}_{j,2}$ respectively, $\sigma$ is a non-linear activation function, and $L$ is the number of hidden layer neurons. As the summation operation is permutation invariant, $L$! equivalent parametrizations are possible for any $\left\{\mathbf{W}_{j, 1},\mathbf{W}_{j, 2}\right\}$ by varying the parameter order. Given the numerous permutations available, the coordinate-wise elements of different local clients’ weight vectors may not necessarily be representing similar feature extraction traits, even for homogeneous local datasets. Thus, the coordinate-wise parameter averaging of the FedAvg algorithm may lead to lower federated model performance due to dissimilar feature extractors (neurons) being averaged and the resultant dampening of feature extraction capabilities. Instead of coordinate-wise parameter averaging, FedMA matches the neurons of local client models based on their similarity first before averaging them to form the federated model neurons. Fig. \ref{fedma_buildup}\subref{fedavg_vs_ma} visually compares the key difference between the FedAvg and FedMA parameter averaging scheme.

The similarity between the local client model neurons as well as the to-be-inferred federated model neurons is formulated by the similarity function $c\left(w_{j l}, \theta_i\right)$, where $w_{j l}$ and $\theta_i$ are the respective weight vectors of client model $j$'s $l$-th neuron and the federated model's $i$-th neuron. To allow for the nonparametric discovery of the federated model neurons, the federated model is framed as a Bayesian probabilistic model that follows the Beta-Bernoulli process (BBP)\cite{wang2020federated}. To compute the maximum a posteriori estimate (MAP) of $\theta_i$, the similarity function $c\left(w_{j l}, \theta_i\right)$ is further defined to be the posterior probability of neuron $l$ from client model $j$ drawn from a Gaussian distribution with mean $\theta_i$. Similarity matching of the client model neurons then requires solving the objective function in Eq. 4 to obtain the required permutations, i.e., neuron matching assignments, $\left\{\pi_{l i}^j\right\}$ that match the indexes of local client model neurons based on their similarity:

\vspace{-0.4cm}
\begin{equation}
\begin{split}
\min _{\left\{\pi_{l i}^j\right\}} \sum_{i=1}^L \sum_{j, l} \min _{\theta_i} \pi_{l i}^j c\left(w_{j l}, \theta_i\right)\hspace{1.4cm} \\
 \text { s.t.} \sum_i \pi_{l i}^j=1 \;\forall\;j, 1 \leq l \leq L_{j} \hspace{0.1cm}; \hspace{0.8cm} \\ \sum_l \pi_{l i}^j=1 \;\forall \; j, 1 \leq i \leq L \hspace{0.1cm}.\hspace{0.9cm}
\end{split}
\vspace{-0.8cm}
\end{equation}
\noindent where, the constraints on $\pi_{li}^j$ are needed to form permutation matrices mathematically.

In practice, it is possible that not all local model neurons can be well matched across clients. Instead of forcing poor matches, the unmatched neuron is added as a new neuron to the federated model. At the same time, excessive federated model growth is limited by a penalty function $f\left(\cdot\right)$ that increases with model size . Thus, to solve for the required permutation $\left\{\pi_{l i}^{j^{\prime}}\right\}_{l, i}$ for a client $j^\prime$ , the objective function in Eq. 4 is expanded to the general form:
\vspace{-0.2cm}
\begin{equation}
\begin{split}
\min _{\left\{\pi_{l i}^{j^{\prime}}\right\}_{l, i}} \hspace{-0.2cm} \mathbf{P} = \hspace{-0.2cm}\sum_{i=1}^{L+L_{j^{\prime}}} \sum_{l=1}^{L_{j^{\prime}}} \pi_{l i}^{j^{\prime}} C_{l i}^{j^{\prime}} \hspace{2.9cm}\\
\text {s.t.} \sum_i \pi_{l i}^{j^{\prime}}=1 \;\forall\; l\hspace{0.1cm};\hspace{3.1cm}\\
\sum_l \pi_{l i}^j \in\{0,1\} \;\forall\; i\hspace{0.1cm},\hspace{2.5cm}\\
\text{where} \hspace{0.1cm} C_{l i}^{j^{\prime}}= \begin{cases}c\left(w_{j^{\prime} l}, \theta_i\right), & i \leq L \\ \epsilon+f(i), & L<i \leq L+L_{j^{\prime}\hspace{0.1cm}.}\end{cases} \hspace{1cm}
\end{split}
\end{equation}
The matching cost threshold $\epsilon$ and penalty function $f\left(\cdot\right)$ are governed by the Indian Buffet Process prior \cite{ghahramani2005infinite}, which models the intuitive phenomenon that if a local neuron is popularly observed in most client models, then there is higher probability for it to be matched and adopted as a global neuron. Further details and proof on the matching cost and penalty formulation are found in \cite{wang2020federated} \cite{yurochkin2019bayesian}.

\begin{figure}[!t]
\centering
\includegraphics[width=0.3\columnwidth]{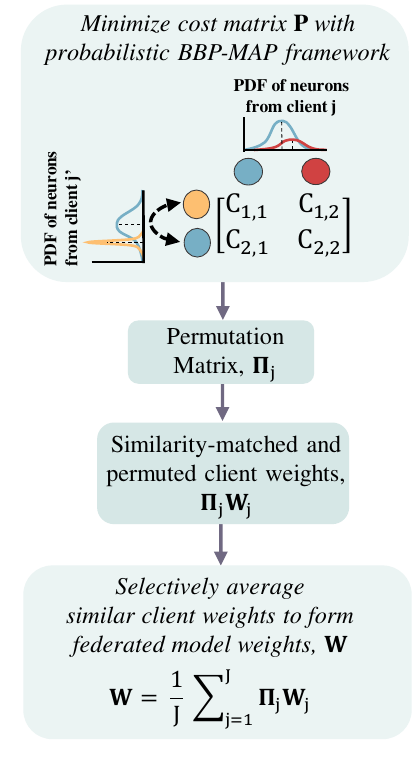}
\vspace{-1.5em}
\caption{\hspace{-0.4em}Similarity-matched averaging scheme for federated model development.}
\vspace{-1.2 em}
\label{breakdown_fedma}
\end{figure}

The optimization problem in Eq. 5 is a linear sum assignment problem \cite{yurochkin2019bayesian}, thus, the Hungarian matching algorithm \cite{kuhn1955hungarian} is applied iteratively over all clients to solve for $\left\{\pi_{l i}^{j^{\prime}}\right\}_{l, i}$, keeping other permutations $\left\{\pi_{l i}^j\right\}_{l, i, j \neq j^{\prime}}$ fixed. For each iteration, the corresponding federated model $\left\{\theta_i=\arg \min _{\theta_i} \sum_{j \neq j^{\prime},l} \pi_{l i}^j c\left(w_{j l}, \theta_i\right)\right\}_{i=1}^L$ is determined, based on current estimates of $\left\{\pi_{l i}^j\right\}_{l, i, j \neq j^{\prime}}$. When faced with a client model $j^{\prime}$ with $L_{j^{\prime}}$ neurons, the federated model is updated by applying the Hungarian algorithm to match the current federated model to candidate local neurons $\left\{w_{j^{\prime} l}\right\}_{l=1}^{L_{j^{\prime}}}$. The new federated model size $L^\prime$ is controlled by the earlier discussed penalty cost $f\left(\cdot\right)$ and limited to $L \leq L^{\prime} \leq L+L_{j^{\prime}}$. This nonparametric federated model framework and the steps to solve Eq. 5 are termed BBP-MAP henceforth.

\subsubsection{Federated Health Prognostic Model Development with FedMA}
Building on the neuron matching formulation, we discuss a layer-wise framework for matching and averaging the client model parameters to construct the $Q$-layer, LSTM-based federated model and determine its parameters. As performing matched averaging recursively through deeper model layers deteriorates the federated model performance \cite{wang2020federated}, a modular, layer-wise approach is adopted, where only the $q$-th layer parameters $\left\{\mathbf{W}_{j,q}\right\}_{j=1}^J$ of the $J$ client models are considered to perform matched averaging and compute the corresponding $q$-th layer parameters $\mathbf{W}_q$ of the federated model. This one-layer matched averaging scheme is repeated for all layers, beginning with all the client models’ first layers (i.e., $q = 1$ in Table \ref{Table_LSTM architecture}) and working down to their last output layers (i.e., $q = Q$). The detailed layer-wise federated model development is outlined below and depicted as a schematic in Fig. \ref{breakdown_fedma}.

Following the local model training described in Section III-A, the clients’  $q$-th layer parameters $\left\{\mathbf{W}_{j,q}\right\}_{j=1}^J$ are gathered by the central server to begin the neuron matching process across clients. The $q$-th layer parameters $\mathbf{W}_{j,q}$ of each client $j$ are transformed to form inputs to Eq. 5 as the concatenated vector of parameters (weights and biases) associated with each of $L_j$ hidden neurons $\left\{w_{j l} \in \mathbb{R}^{UD_{in}+U}\right\}_{l=1}^{L_j}$, where $U$ is the number of intermediate LSTM cell weights (i.e., 4)\cite{hochreiter1997long} and $D_{in}$ is the input size. These weight vectors are iteratively fed into the probabilistic BBP-MAP cost formulation discussed in Section III-B(1) to solve Eq. 5 for the required index assignments $\left\{\pi_{li}^j\right\}$, rewritten as $\mathbf{\Pi}_j^T$, that match client $j$’s neurons to similar neurons of all other clients excluding $j$.

\begin{algorithm}[t]
\caption{\small FedMA-based Health Prognostic Model Development}
\footnotesize
\algsetup{linenosize=\footnotesize}
\begin{algorithmic}[1]
\renewcommand{\algorithmicrequire}{\textbf{Input:}}
\renewcommand{\algorithmicensure}{\textbf{Output:}}
\newcommand{\algorithmicbreak}{\textbf{break}}
\newcommand{\BREAK}{\STATE \algorithmicbreak}
\REQUIRE Local parameters $\left\{\mathbf{W}_{j, 1}, \ldots, \mathbf{W}_{j, Q}\right\}_{j=1}^J$ of $J$ client models with \\
\hspace{3.6mm} a $Q$-layer architecture
\ENSURE Federated model parameters $\left\{\mathbf{W}_1, \ldots, \mathbf{W}_Q\right\}$
\\ \text{Initialize} layer index $q = 1$
\WHILE {($q \leq Q$)}
 \IF {($q < Q$)}
 \STATE $\left\{\mathbf{\Pi}_j\right\}_{j=1}^J=\operatorname{BBP-MAP}\left(\left\{\mathbf{W}_{j, q}\right\}_{j=1}^J\right)$;
 \STATE $\mathbf{W}_q=\frac{1}{J} \sum_j \mathbf{W}_{j, q} \mathbf{\Pi}_j^T$;
 \ELSE
 \STATE $\mathbf{W}_q=\frac{1}{J} \sum_j \mathbf{W}_{j, q}$;
 \ENDIF
 \FOR {$j \in\{1, \ldots, J\}$}
 \STATE $\mathbf{W}_{j, q} \leftarrow \mathbf{W}_q$;
 \STATE $\mathbf{W}_{j, q+1} \leftarrow \mathbf{\Pi}_j \mathbf{W}_{j, q+1}$;
 \STATE Retrain $\left\{\mathbf{W}_{j, q+1}, \ldots, \mathbf{W}_{j, Q}\right\}$ with $\mathbf{W}_q$ frozen;
 \ENDFOR
 \STATE $q=q+1$
\ENDWHILE
\end{algorithmic}
\end{algorithm}

Within each LSTM layer, there are both the input-to-hidden and hidden-to-hidden components as shown in Table \ref{Table_LSTM architecture}. For input-to-hidden weights $\mathbf{W}_{j,q}$, the matched index assignments $\mathbf{\Pi}_j^T$ are multiplied to the right (acts on columns) of the original client weights $\mathbf{W}_{j,q}$ to yield the correctly permuted client weights $\mathbf{W}_{j,q}\mathbf{\Pi}_{j }^T$, where similar neurons now hold the same index position across different clients’ weight matrices. The weights $\mathbf{W}_q$ for the federated model’s $q$-th layer is then computed by averaging the similarity-matched $q$-th layer parameters of the $J$ client models:
\begin{equation}
\mathbf{W}_q=\frac{1}{J} \sum_{j=1}^J \mathbf{W}_{j, q} \boldsymbol{\Pi}_j^T
\end{equation}
For the hidden-to-hidden parameters $\mathbf{H}_{j,q}$, both the rows and columns of $\mathbf{H}_{j,q}$ are permutable, thus, the solved permutation matrix $\mathbf{\Pi}_j$ is applied to the left and right of $\mathbf{H}_{j,q}$. The hidden-to-hidden parameters $\mathbf{H}_q$ for the $q$-th layer of the federated model is then computed by averaging the similarity-matched hidden layer weights of the $J$ client models:
\begin{equation}
\mathbf{H}_q=\frac{1}{J} \sum_{j=1}^J \boldsymbol{\Pi}_j \mathbf{H}_{j, q} \boldsymbol{\Pi}_j^T
\end{equation}

In the current layer-wise approach, once the $q$-th layer federated parameters $\mathbf{W}_q$ (includes both input-to-hidden and hidden-to-hidden parameters for simplicity) are computed, it is distributed back to the local client models first, to be initialized as their respective $q$-th layer parameters. Keeping their initialized $q$-th layers frozen, clients retrain all subsequent layers of their models on their private datasets to generate updated parameters for the remaining layers based on the $q$-th layer initialization. This process of conducting similarity-based matched averaging on a selected $q$-th layer of client models to determine the federated model layer, followed by local client model retraining of subsequent layers is continued until the last output layer. For the last layer, there is no ambiguity in the ordering of the output nodes (the permutation matrix is, in fact, an identity matrix). Thus, a simple mean of the last layer local model parameters yields the federated model’s last layer parameters. The complete FedMA round to determine all parameters $\mathbf{W}=\left\{\mathbf{W}_{q}\right\}_{q=1}^{Q}$ of the $Q$-layer federated model is summarized in Algorithm 1. Once the first iteration of the federated model is constructed, the model performance can be further improved by repeating the FedMA procedure for more communication rounds between the server and client models until a desired accuracy or model convergence is achieved.

\subsection{Online Health Prognosis}
Once the federated model is sufficiently trained offline, it is deployed back to local clients, which now have a health prognosis model that has been exposed to a wider range of operational and machine failure patterns despite local data limitations and privacy constraints. With the federated model, clients can individually conduct online inferencing tasks on incoming private test data $\mathbf{X}_{{test}}$.  For SoH estimation with cyclic test data and RUL estimation with non-cyclic test data, the pre-processing transformations outlined in the respective Sections, III-A(1) and  III-A(2) are applied to $\mathbf{X}_{test}$ prior to inputting it into the LSTM-based federated model, $\widehat{Y}_{test}=\phi\left(\mathbf{X}_{test}, \mathbf{W}\right)$. Here, $\phi$ is a general, non-linear representation of either an SoH estimation or RUL estimation model, $\mathbf{W}$ is the trained federated model parameters from Section III-B(2), and $\widehat{Y}_{test}$ is the predicted SoH or RUL value.

\begin{table}[!t]
\newcolumntype{P}[1]{>{\centering\arraybackslash}p{#1}}
\renewcommand{\arraystretch}{1.3}
\scriptsize
\centering
\caption{\centering Experimental settings to assess robustness of FL techniques.}
\vspace{-2 em}
\label{table1_exp settings}
\begin{center} 
\begin{tabular}{P{3.5cm}P{4.2cm}}
\hline
Experiment 1: \hspace{30mm} SoH estimation of batteries & Experiment 2: \hspace{30mm} RUL estimation of turbofan engines \\
\hline
\vspace{-2mm}
\begin{itemize}
\item Cyclic degradation data \vspace{0.5mm}
\item Cross-device FL within the same organization \vspace{0.5mm}
\item Heterogeneous local devices with dissimilar degradation processes
\end{itemize}
&
\vspace{-2mm}
\begin{itemize}
  \item Non-cyclic degradation data \vspace{0.1mm}
  \item Cross-organization FL where each organization owns multiple devices\vspace{0.5mm}
  \item Heterogeneous organizations with dissimilar degradation processes among groups of devices and unbalanced training datasets
\end{itemize}\\
\hline
\vspace{-1.2 em}
\end{tabular}
\end{center}
\vspace{-2 em}
\end{table}

\section{Experiments and Results}
This section evaluates the SoH and RUL estimation performance of FedMA-based models against the standard FedAvg-based models and importantly, the baseline local models without FL.  Using two distinctly different datasets (Li-ion batteries and turbofan engines), the benchmarkable  experiments have been thoughtfully designed to assess the generalizability and robustness of the studied FL methodologies in a variety of settings such as i) cyclic vs. non-cyclic degradation data, ii) cross-device FL within the same organization vs. cross-organization FL, and iii) device-level heterogeneity vs. organization-level heterogeneity. Table \ref{table1_exp settings} summarizes the settings examined for each experiment. In addition to FL performance analysis, each experiment also discusses practical insights, such as the potential for training data efficiency in Experiment 1 and the interpretability of similar feature extraction traits in Experiment 2.
\begin{table}[!t]
\newcolumntype{P}[1]{>{\centering\arraybackslash}p{#1}}
\renewcommand{\arraystretch}{1.2}
\centering
\scriptsize
\caption{\centering Description of NASA battery dataset.}
\vspace{-2em}
\label{table2_battery description}
\begin{center} 
\begin{tabular}{P{0.7cm}|P{0.8cm}P{2cm}P{1.6cm}P{1.3cm}}
\hline
Battery & Cut-off voltage & Battery life cycles before EoL & Training cycles $(70 \%$ split) & Test cycles $(30 \%$ split) \\
\hline B0006 & $2.5 \mathrm{~V}$ & 121 & 85 & 36 \\
B0007 & $2.2 \mathrm{~V}$ & 168 & 118 & 50 \\
B0018 & $2.5 \mathrm{~V}$ & 122 & 85 & 37 \\
\hline
\end{tabular}
\end{center}
\vspace{-1.2 em}
\end{table}

\begin{figure}[!t]
\vspace{-1.2 em}
\centering
\subfloat[]{\includegraphics[width=0.4\linewidth]{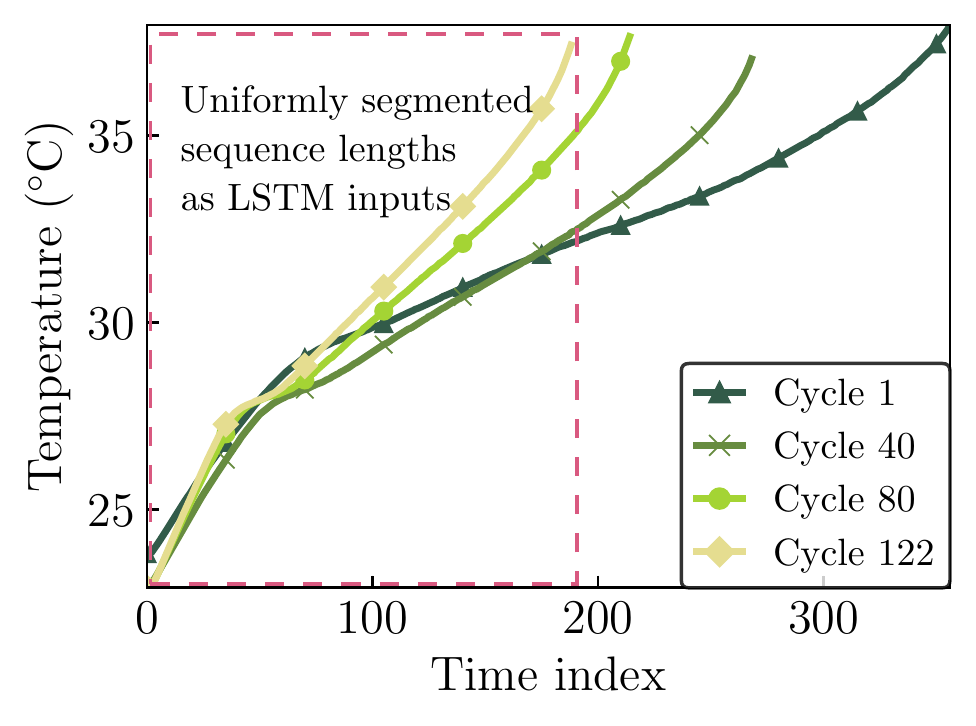}%
\label{temperature}}
\hfil
\subfloat[]{\includegraphics[width=0.4\linewidth]{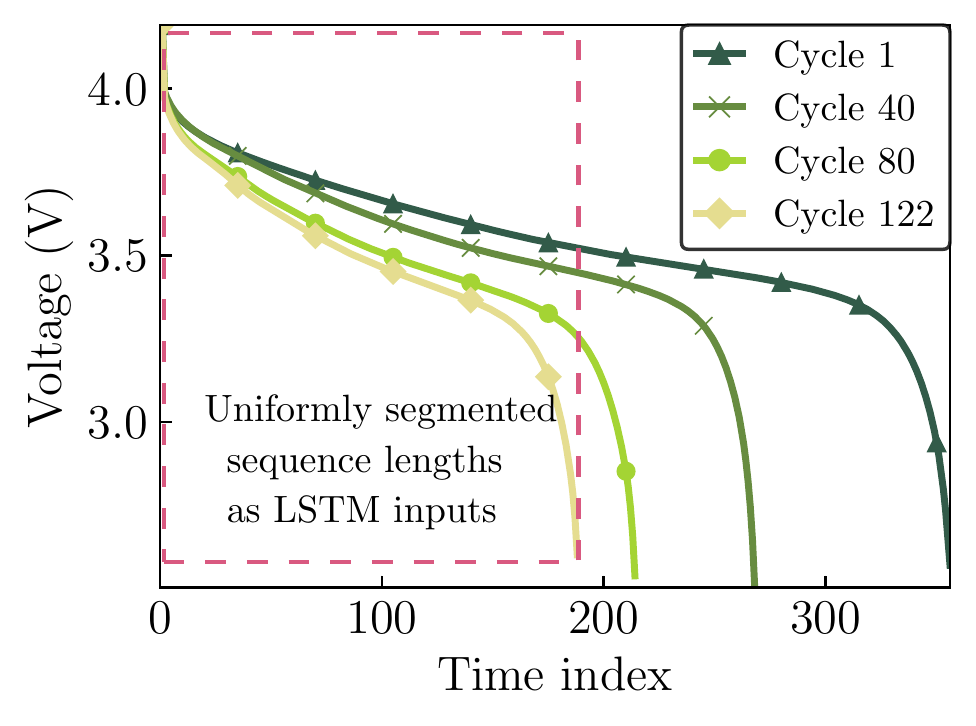}%
\label{voltage}}
\caption{(a) Temperature signals, and (b) voltage signals of the battery B0018 from the NASA dataset.}
\vspace{-1.2em}
\label{temp_voltage}
\end{figure}

\subsection{Experiment 1: SoH Estimation of Li-ion Batteries}
In this problem formulation, each battery is a client requiring an accurate, on-device SoH estimation model. Under privacy constraints, clients have to develop their local models independently using only their respective datasets for model training. However, with FL, a richer federated model can be built through collaborative learning from multiple client models and utilized back by the clients. As the batteries are all of the same type, this experiment is akin to a real-world, cross-device FL setup within a single organization. The accuracy of the resultant federated model is evaluated.

\subsubsection{Data Description}
The NASA battery dataset\cite{saha2007battery} consists of 2 Ah-rated Li-ion cells that have undergone repeated charging and discharging cycles in accelerated ageing experiments. Conventionally, it is the discharging cycle that is relevant to SoH estimation as the discharging voltage, current, and temperature measurements form the input features for the capacity estimation model. The batteries were discharged from a voltage of 4.2V to a cut-off voltage of either 2.2V or 2.5V at a constant 2A current and an ambient temperature of 24°C (though the actual temperature of the battery varies as it heats up during the charge-discharge operations). Taking battery B0018 as an example, Fig. \ref{temp_voltage} illustrates the progression of the measured voltage and temperature time series within each discharge cycle for selected discharge cycles. Generally, the length of a discharge cycle, given by the length of the time index, decreases from the first cycle to the last.

\begin{figure}[!t]
\centering
\includegraphics[width=0.5\linewidth]{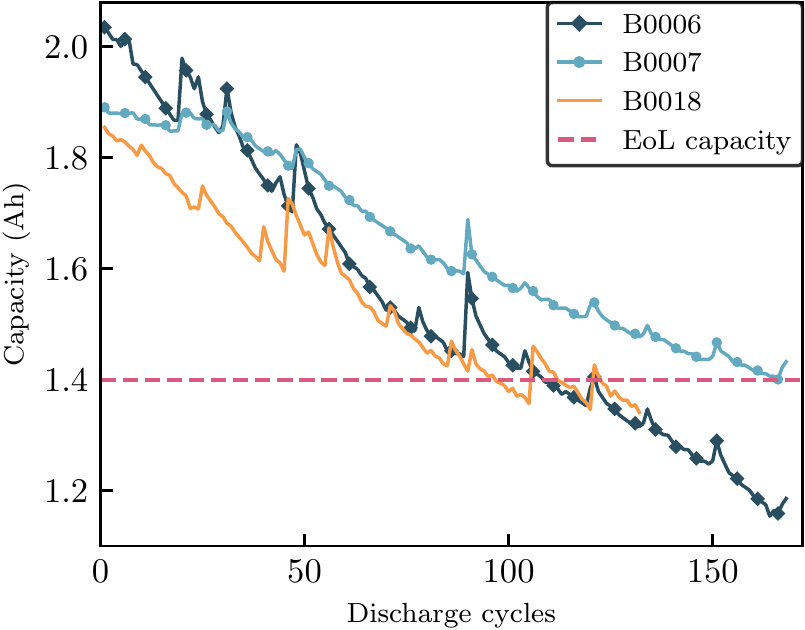}
\vspace{-1em}
\caption{Capacity loss over cycles for batteries, ‘B0006’, ‘B0007’, and ‘B0018’.}
\vspace{-1.2 em}
\label{capacityfade}
\end{figure}

As the discharging cycles progress, capacity degradation also occurs until the end-of-life (EoL) capacity, defined as 70\% of original rated capacity. Only battery data up till EoL are used for SoH estimation modeling, with the first 70\% of cycles allocated for model training and the remaining 30\% for testing. The salient attributes of the dataset are summarized in Table \ref{table2_battery description}. To examine the battery degradation patterns further, the capacity loss over the discharge cycles is plotted for each battery in Fig. \ref{capacityfade}. It is evident that the capacity degradation follows a cyclic pattern—amidst the overall decreasing trend, there exists periodic localized peaks due to the capacity regeneration phenomenon \cite{ma2021remaining}. Secondly, the three batteries exhibiting notably different capacity degradation pathways despite having similar discharging conditions reiterates the need for FL techniques that account for local data heterogeneity.
\begin{table}{}
\newcolumntype{P}[1]{>{\centering\arraybackslash}p{#1}}
\renewcommand{\arraystretch}{1.3}
\scriptsize
\centering
\caption{Hyperparameter selection for local model architecture.}
\label{Table3_hyperparameter}
\begin{threeparttable}
\begin{tabular}{p{1.8cm}P{2.3cm}P{1.6cm}P{1.6cm}}
\hline Hyperparameters & Search space & Selected value (Experiment 1) & Selected value (Experiment 2) \\
\hline Sequence length & $\{171\}$, $\{30,40,50\}$ & $171\tnote{a}$ & 50 \\
LSTM layers & $\{1,2\}$ & 1 & 1 \\
Neurons per layer & $\{50,128,256,300\}$ & 128 & 256 \\
Learning rate & $\{0.0005,0.001,0.005\}$ & $0.001$ & $0.001$ \\
Optimizer & RMSProp, Adam & Adam & Adam \\
\hline
\end{tabular}
\begin{tablenotes}
\item[a] Sequence length is determined by the length of the shortest discharge cycle across all batteries.
\end{tablenotes}
\end{threeparttable}
\vspace{-1.2 em}
\end{table}

\begin{table*}
\newcolumntype{P}[1]{>{\centering\arraybackslash}p{#1}}
\renewcommand{\arraystretch}{1.3}
\scriptsize
\centering
\caption{Performance comparison of FL models against locally trained models and centrally trained models with pooled data}
\label{Table4_battery results}
\begin{threeparttable}
\begin{tabular}{P{2cm}P{2.4cm}P{2cm}P{1.4cm}P{2.7cm}P{1.2cm}}
\hline Test Client Battery & Train Client Batteries & Method & RMSE & Best communication round & IMP\tnote{a} \\
\hline \multirow{5}{*}{B0006}& B0006 & Local data only & 0.04979 & - & - \\
& B0006, B0007, B0018 & Central & 0.05307 & - & - \\
& B0006, B0007, B0018 & FedAvg & 0.04588 & 4 & 7.9 \% \\
& B0006, B0007, B0018 & FedMA \textbf{(proposed)} & \textbf{0.04381} & \textbf{15} & \textbf{12.0 \%} \\
& B0006, B0018 & FedMA \textbf{(proposed)} & \textbf{0.03871} & \textbf{4} & \textbf{22.3 \%} \\
\hline

\multirow{5}{*}{B0007}& B0007 & Local data only & 0.03610 & - & - \\
& B0006, B0007, B0018 & Central & 0.03296 & - & - \\
& B0006, B0007, B0018 & FedAvg & 0.02933 & 4 & 18.8 \% \\
& B0006, B0007, B0018 & FedMA \textbf{(proposed)} & \textbf{0.02502} & \textbf{16} & \textbf{30.7 \%} \\
& B0006, B0007 & FedMA \textbf{(proposed)} & \textbf{0.02193} & \textbf{5} & \textbf{39.2 \%} \\
\hline
\multirow{5}{*}{B0018} & B0018 & Local data only & 0.03361 & - & - \\
& B0006, B0007, B0018 & Central & 0.02972 & - & - \\
& B0006, B0007, B0018 & FedAvg & 0.02868 & 4 & 14.7 \% \\
& B0006, B0007, B0018 & FedMA \textbf{(proposed)} & \textbf{0.02257} & \textbf{16} & \textbf{32.9 \%} \\
& B0006, B0018 & FedMA \textbf{(proposed)} & \textbf{0.01865} & \textbf{14} & \textbf{44.5 \%} \\

\hline
\end{tabular}
\begin{tablenotes}
\item[a] IMP is short for improvement, which is defined as (‘Local data only’ \textit{RMSE} – FL \textit{RMSE}) / ‘Local data only’ \textit{RMSE}.
\end{tablenotes}
\end{threeparttable}
\vspace{-1.2 em}
\end{table*}

\begin{figure*}[!t]
\centering
\subfloat[B0006]{\includegraphics[width=0.32\linewidth]{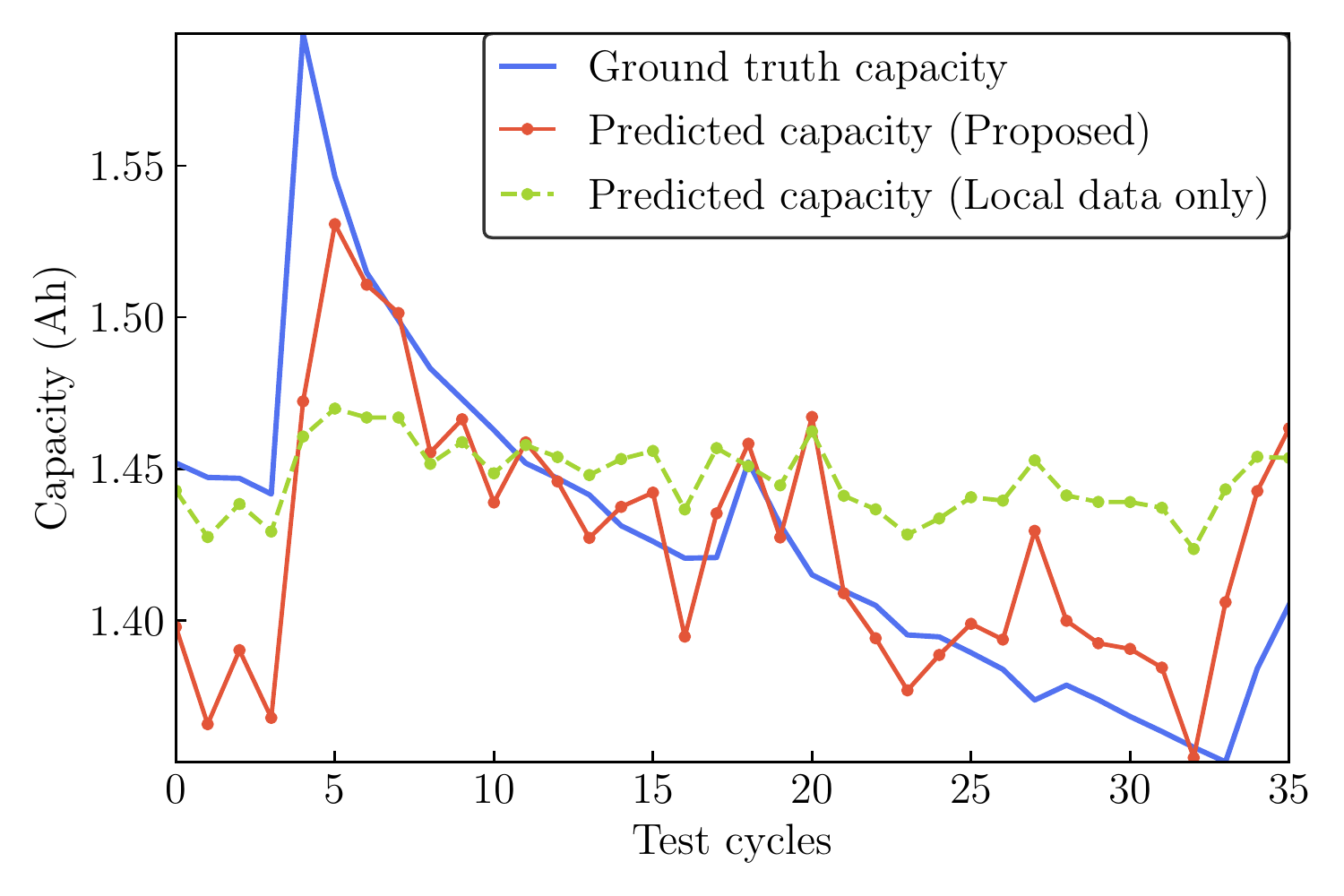}%
\label{B0006}}
\hfil
\quad
\subfloat[B0007]{\includegraphics[width=0.32\linewidth]{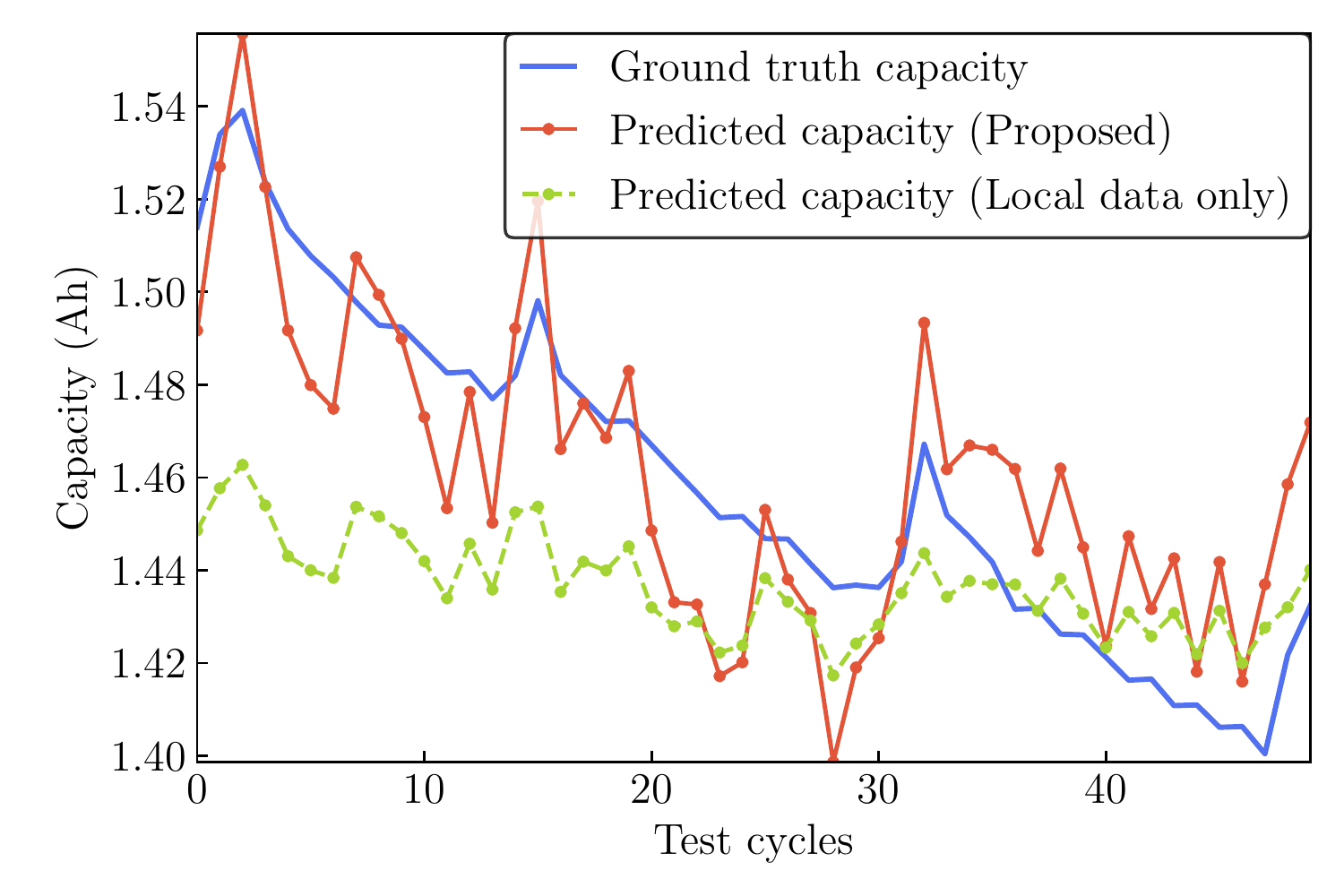}%
\label{B0007}}
\hfil
\quad
\subfloat[B0018]{\includegraphics[width=0.32\linewidth]{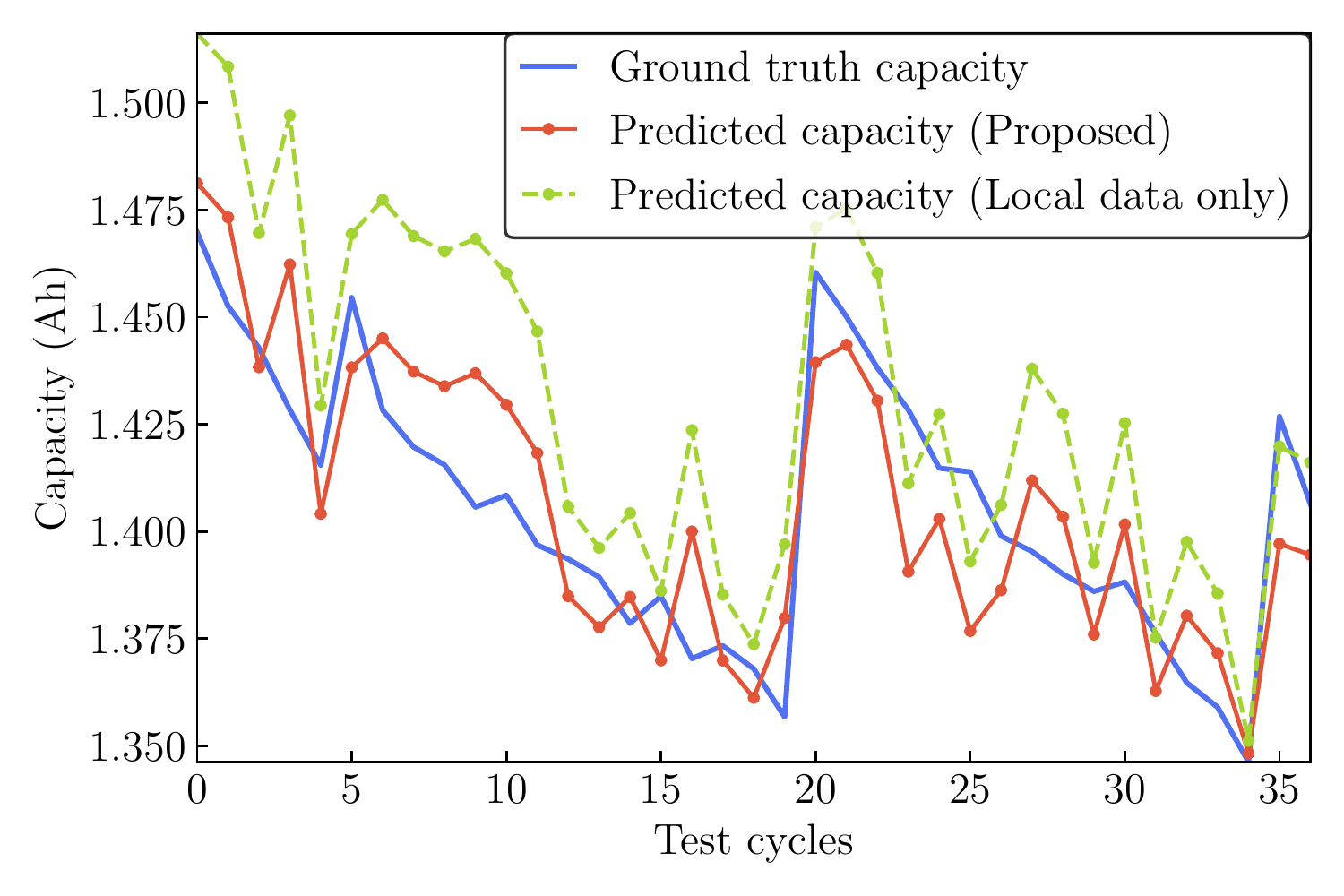}%
\label{B0018}}
\caption{Predictive performance of local model without any FL collaboration (local data only) vs. proposed approach for test cycles from batteries (a) B0006, (b) B0007, and (c) B0018.}
\vspace{-1.3em}
\label{battery performance}
\end{figure*}

\subsubsection{Local Client Model Formulation for SoH Estimation}
First, the local model for each client battery is developed. Using only the first 70\% of a battery’s own discharge cycles as training data, a LSTM-based SoH estimation framework is built to model the non-linear relationship between the feature variables, discharging voltage and temperature, and the capacity labels. The discharging current is not selected as a feature as the batteries were discharged at constant current and thus, the current measurements do not provide useful information about the capacity degradation process. Two key pre-processing steps are performed on the features to transform them as suitable LSTM inputs. Features are standardized to have zero mean and unit variance so that large values do not dominate and skew model training and convergence. As the LSTM model requires fixed length input sequences, the standardized feature time series are then segmented into sequences of a common length determined by the shortest discharge cycle as depicted in Fig. \ref{temp_voltage}. These feature sequences, together with their corresponding capacity labels, are consecutively fed into the LSTM framework to model the progression of capacity levels through the training discharge cycles and estimate the capacity, or equivalently, SoH for each battery's test data (i.e., the earlier partitioned last 30\% of discharge cycles). For the local model architecture, a lighter model is preferred for a lower communication burden between the server and the clients during FL. Keeping this in mind, the model variations listed in Table \ref{Table3_hyperparameter} were assessed. All models were trained for 100 epochs with early stopping. A one-layer LSTM network with 128 neurons, followed by a fully connected output layer (therefore, $Q=2$), emerged as the best-performing SoH estimation model for the test data. Each client’s best local model accuracy represents the maximum achievable performance by each client under privacy constraints without any FL and it forms the baseline for comparing the FL performance later in Section III-A(4).

\subsubsection{Federated Model Development for SoH Estimation}
Using the same 2-layer architecture as the local models, the federated models are developed next using the FedAvg and FedMA algorithms. The FedAvg-based federated model is constructed following the steps in Section II-C, where the local model parameters are aggregated coordinate-wise using Eq. 2. According to the FedAvg algorithm, the retraining of local models following their initialization with the federated model parameters has to be performed incrementally for $E$ epochs every communication round to reduce the possibility of FL training overshooting and diverging \cite{wang2020federated}. As the local models previously needed about 100 epochs for convergence in the baseline scenario without FL, $E \in\{2,5,10\}$ were considered as candidate epochs for the incremental training. The best federated model performance was achieved with $E=2$ for local model retraining.
In contrast, the FedMA algorithm favors longer retraining and convergence of local models before the local model parameters are aggregated. Sufficiently converged local models ensure that the similarity matching of neurons are performed based on stable, high quality feature extraction traits. The FedMA-based federated model is constructed following Algorithm 1, with local model retraining conducted for 120 epochs within each communication round. Compared to the 100 epochs used for local model training in the baseline scenario without FL, the local model retraining for FedMA is understandably slightly longer since the collaborative learning among multiple clients exposes more degradation patterns to be learnt.

\subsubsection{SoH Estimation Results and Discussion}
The FedMA and FedAvg federated models were both trained up to 20 communication rounds. For each round, the \textit{RMSE} of the capacity predictions was computed to evaluate the federated model accuracy on local clients’ test data. The best federated model accuracy, the number of communication rounds needed, and the accuracy improvement over baseline client models without any FL are presented in Table \ref{Table4_battery results}. Generally, the federated models outperformed client models trained on only their own data, which reiterates the crucial role FL plays in allowing local client models to benefit from a wider range of training data despite privacy constraints. Comparing the two FL methods, FedMA achieved at least 1.5 times  larger accuracy improvements than FedAvg for all client batteries. While FedAvg needed less communication rounds to reach its best model, it also converged to sub-optimal results than FedMA due to its coordinate-wise parameter aggregation algorithm, which ignores local data heterogeneity.

Against the known premise that more training data leads to better model performance, the scope for \textbf{data-efficiency} in  FL was also explored since utilizing less clients offers practical advantages such as reduced communication burden and training time. For FedMA models, training on a smaller select subset of clients was viable, as shown in Table IV. Not only did the federated model accuracy significantly increase (lowest \textit{RMSE}), the communication rounds required to achieve the best model also decreased. In contrast, FedAvg models generally worsened when trained on smaller client subsets (the additional FedAvg results are not reported for brevity). A possible reason for FedMA’s notable outperformance in this data-efficient scenario is that the beneficial presence of certain compatible clients is amplified by the matching of similar client neurons. The resultant aggregated federated model has sharper feature extraction capabilities and hence, accuracy. For an in-depth look into the SoH estimation performance of the FedMA federated model, Fig. \ref{battery performance} plots the predicted capacities against the ground truth labels for each client battery’s test data. Compared to the baseline client models trained on only their limited data, FedMA’s predictions exhibit remarkable closeness to ground truth capacities and their trend.

As an added evaluation criterion, the FL models are compared against the ideal setting, where there are no privacy constraints and edge data can be directly pooled for central training. This central training scenario should theoretically have the best model performance (and forms the upper bound for FL performance) because it is able to directly utilize all client data instead of relying on model parameter aggregation. As expected, the ‘Central’ models, reported in Table \ref{Table4_battery results}, outperformed most client batteries’ local models trained only on local data. However, contrary to initial expectations, the ‘Central’ models consistently underperformed FL models, revealing a noteworthy advantage of FL. The underperformance of the ‘Central’ models in this experiment is likely due to the cyclic nature of battery capacity degradation data (as previously seen in Fig. \ref{capacityfade}). As opposed to non-cyclic, monotonously decreasing data, centrally aggregating cyclic data of different batteries challenges the model’s ability to distinguish between regular cyclic patterns of the current battery and the start of the next battery’s discharge cycles. In contrast, FL models consider each battery’s model and indirectly, the data separately. Thus, they can reap the benefit of learning from multiple client batteries without being disturbed by cyclic noise.

\subsection{Experiment 2: RUL Estimation of Turbofan Engines}
The second experiment is designed to investigate the scalability of FL methods in a cross-organization setting, where each organization (client) owns a heterogeneous sample of training devices but needs to conduct RUL estimation for some test devices beyond the distribution  of their training samples. Heterogeneity, in the form of unequal amount of training data samples and devices with dissimilar degradation processes, is introduced to clients. The RUL estimation performance of the FedMA and FedAvg models amidst the data heterogeneity  is assessed.

\begin{table*}
\newcolumntype{P}[1]{>{\centering\arraybackslash}p{#1}}
\renewcommand{\arraystretch}{1.3}
\scriptsize
\centering
\caption{Performance comparison of FL models against local models and centrally trained models with pooled data.}
\label{FL_cmapps_results}
\begin{threeparttable}
\begin{tabular}{P{1.5cm}P{2.2cm}P{1.3cm}P{1.1cm}P{1.7cm}P{1cm}P{1cm}P{0.005cm}P{1cm}P{1.2cm}P{1cm}}
\hline
\multirow{3}{*}{Setting} & \multirow{3}{*}{Clients} & \multirow{3}{1.3cm}{\centering No. of train engines} & \multirow{3}{1.1cm}{\centering No. of test engines\tnote{a}} &  \multicolumn{3}{c}{\centering \textit{RMSE}} & & \multicolumn{2}{c}{Best communication round}
& \multirow{3}{1cm}{\centering IMP\tnote{b}}\\
\cline{5-7}\cline{9-10}
& & & & Local data only/ Central pooled & \multirow{2}{1cm}{\centering FedAvg} & \multirow{2}{1cm}{\centering FedMA \textbf{(proposed)}} & & \multirow{2}{1cm}{\centering FedAvg} & \multirow{2}{1.2cm}{\centering FedMA \textbf{(proposed)}} \\
\hline
\multirow{3}{1.5cm}{\centering Heterogeneous partitioning} & HT1
($<$200 cycles) & 82 & 228 & 39.26 &\multirow{3}{*}{28.61}& \multirow{3}{*}{\textbf{27.22}} & & \multirow{3}{*}{9}& \multirow{3}{*}{\textbf{4}} & \textbf{30.7\%}\\
\cline{2-5} \cline{11-11}

& HT2 (200-350 cycles) & 25 & 228 & 44.87 & & & & & & \textbf{1.9\%}\\
\cline{2-5} \cline{11-11}
& HT3 ($>$ 350 cycles) & 142 & 228 & 27.76 & & & & & & \textbf{39.3\%}\\
\hline
\multirow{3}{1.5cm}{\centering Homogeneous partitioning} & HO1 & 83 & 228 & 32.94 &\multirow{3}{*}{30.42}& \multirow{3}{*}{\textbf{30.09}} & & \multirow{3}{*}{13}& \multirow{3}{*}{\textbf{5}} & \textbf{8.7\%}\\
\cline{2-5} \cline{11-11}

& HO2 & 83 & 228 & 32.27 & & & & & & \textbf{6.8\%}\\
\cline{2-5} \cline{11-11}

& HO3 & 83 & 228 & 36.15 & & & & & & \textbf{16.8\%}\\
\hline
Central & NA\tnote{c} & 249 & 228 & 26.75 & NA\tnote{c} & NA\tnote{c} &  NA\tnote{c} & NA\tnote{c} & NA\tnote{c} & -1.8\%\\
\hline
\end{tabular}
\begin{tablenotes}
\item[a] Although there are 248 test engines in the FD004 dataset, 20 engines were excluded as their data provided is insufficient for the LSTM model sequence length of 50.
\item[b] IMP is short for improvement, which is defined as (‘Local data only’ \textit{RMSE} – FedMA \textit{RMSE})/‘Local data only’ \textit{RMSE}
\item[c] NA is short for not applicable.
\end{tablenotes}
\end{threeparttable}
\vspace{-1.2 em}
\end{table*}

\subsubsection{Data Description}
The Commercial Modular Aero-Propulsion System Simulation (C-MAPSS) turbofan engine degradation dataset \cite{saxena2008damage} contains simulation results from run-to-failure experiments of multiple engines. The dataset is a collection of 4 sub-datasets, FD001 to FD004, which vary in the number of engines, fault types, and operating conditions studied. For each sub-dataset, predefined sets of train and test engines are also provided. Since FD004 is the most complex sub-dataset with engines experiencing 6 operating conditions and 2 fault types, it is selected for our FL experiment, which focuses on client data heterogeneity.

\begin{figure}[!t]
\vspace{-1.2 em}
\centering
\subfloat[]{\includegraphics[width=0.4\linewidth]{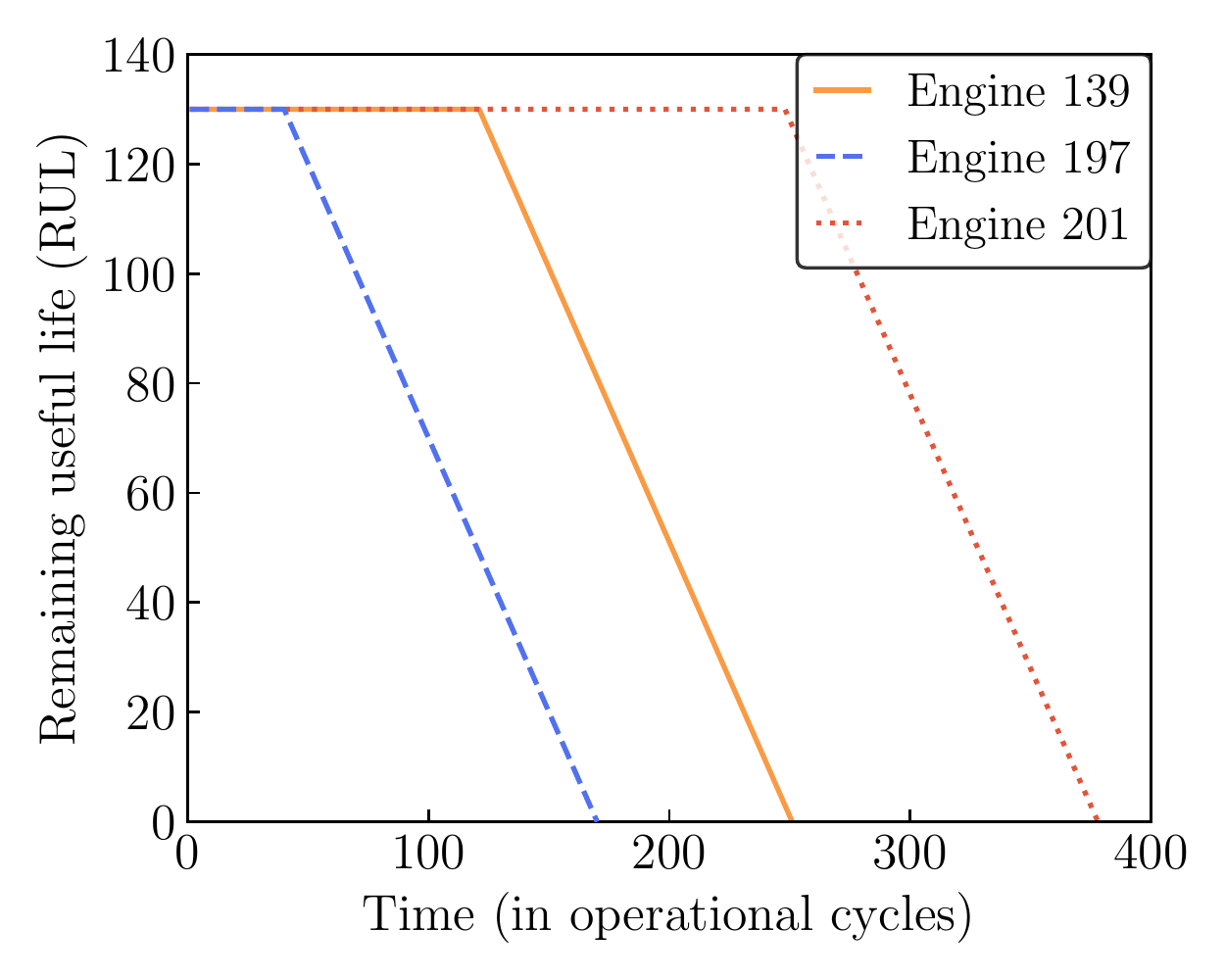}%
\label{piecewise function}}
\hfil
\subfloat[]{\includegraphics[width=0.4\linewidth]{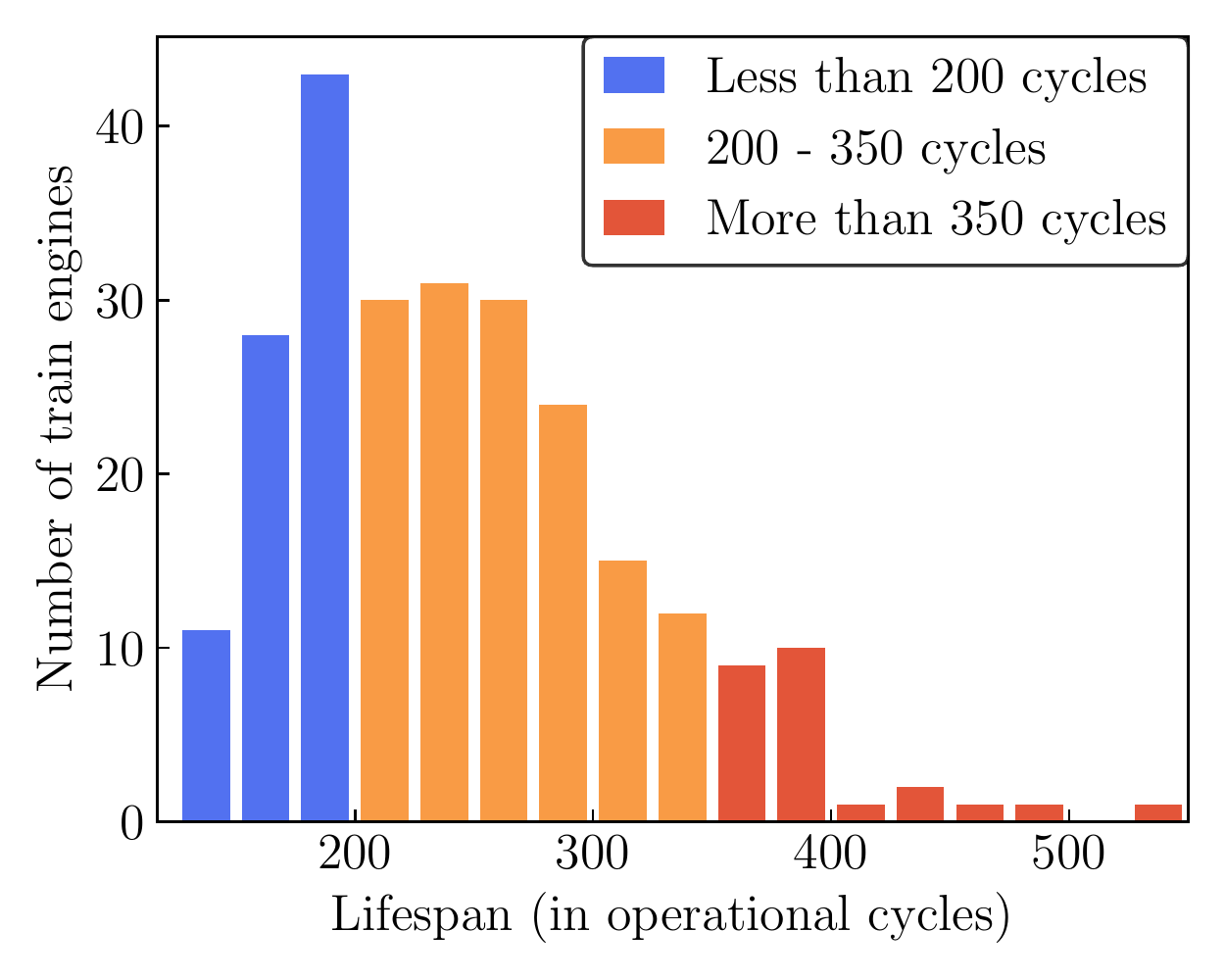}%
\label{lifespan distribution}}
\caption{(a) Piecewise RUL representation of selected train engines (b) distribution of train engine lifespans.}
\vspace{-1.6em}
\label{cmapss_description}
\end{figure}

In the FD004 dataset, measurements for 21 sensor variables (e.g., pressure, temperature) are given for each operational time cycle of an engine. Following [10], sensors 1, 5, 6, 10, 16, 18, and 19 are not utilized for model training as they are either constant with time or exhibit erratic trends. The remaining sensors form the input features for the RUL estimation model. The RUL labels at each operational cycle of the train engines are not provided, but they can be constructed by taking the difference between the lifespan of the engine and the current operational cycle. Generally, it is more realistic to represent the progression of RUL as a non-linear, piecewise function because an engine’s RUL remains fairly constant initially and only decreases after some time in operation when degradation begins. The empirical  value of 130\cite{zheng2017long}, commonly used in turbofan engine literature, is used as the initial constant value for the RUL. The RUL labels of a few train engines are plotted in Fig. \ref{cmapss_description}\subref{piecewise function} to illustrate the piecewise representation. It can be seen that turbofan engines have non-cyclic and monotonous degradation patterns.

Next, the train engines are partitioned to form clients for FL training. Though the engines are of the same type, their lifespans can vary considerably depending on the operating conditions. Noting that the lifespans of train engines range from 128 to 543 operating cycles in Fig. \ref{cmapss_description}\subref{lifespan distribution}, the train engines are divided into the following three categories:
\begin{itemize}
  \item Short-life: Lifespan less than 200 cycles
  \item Moderate-life: Lifespan between 200 to 350 cycles
  \item Long-life: Lifespan more than 350 cycles
\end{itemize}
For the heterogeneous client formulation, a practical setting is envisaged, where clients have considerably different engine usage and data collection schedules. Thus, each client can end up with training data from only a particular category of lifespan distribution. These heterogeneous clients, HT1 to HT3, are detailed in the first three columns of Table \ref{FL_cmapps_results} The clients differ in both the engine lifespan distribution as well as the engine dataset sizes owned. For instance, while client HT3 has data on longest lasting engines (more than 350 cycles), it also has the least amount of data (25 engines). To juxtapose the heterogeneous setting, a homogeneous partitioning of the client is also considered. The entire train engine dataset is randomly shuffled to form the clients, HO1 to HO3, as detailed in Table \ref{FL_cmapps_results}. Each client has engines with a lifespan distribution similar to the global train engine dataset and an equal amount of data (83 engines each).

\subsubsection{Local Client Model Formulation for RUL Estimation}
The FD004 test engines have the same feature variables as the train engines, but the data are only provided up to some random time before failure for each engine. Thus, the remaining operating cycles before failure, i.e., RUL, is not known. The problem setup is that the local clients are tasked with estimating the RUL of the test engines, which may have a different lifespan distribution from the client’s available training data. Using each client’s train engine data, an LSTM-based RUL estimation model relating the sensor data and the piecewise RUL labels is developed for the baseline  client models.
The best model architecture for the client models, which also forms the basis for the federated models, is determined in a manner broadly similar to Experiment 1. The sensor data are standardized first (zero mean and unit variance) and then fed into the LSTM model as smaller, uniform-length input sequences for model training. The model variations in Table \ref{Table3_hyperparameter} were evaluated. All models were trained for a maximum of 300 epochs with early stopping. A one-layer LSTM network with 256 neurons, encompassed by a fully connected input and output layer (therefore, $Q=2$), emerged as the best-performing RUL estimation model when assessed on the test engines.

\subsubsection{Federated Model Development for RUL Estimation}
Using the same 2-layer architecture, the federated models are developed next using the FedAvg and FedMA algorithms.  For the FedAvg federated model development, recall from Section IV-A(2) that the local model retraining within each communication round has to be performed incrementally for $E$ epochs per round to curtail FL training divergence. Since about 300 epochs were required for local model convergence in the baseline scenario without FL, a set of small and mid-ranged values of $E \in\{5,10,15,50,75\}$ were considered as candidate epochs.  The best FedAvg federated model performance was achieved with $E=15$ for heterogeneous clients and $E=10$ for homogeneous clients. The FedMA model development, in contrast, favors longer local model retraining within each communication round as explained earlier in Section IV-A(3). Accordingly, local model retraining was conducted for a maximum of 200 epochs within each communication round.

\subsubsection{RUL Estimation Results and Discussion}
The FedMA and FedAvg federated models were trained up to 15 communication rounds for both the heterogeneous and homogeneous client settings. For each round, the federated model accuracy is assessed through the \textit{RMSE} of the RUL predictions on the test engines. The results are comprehensively presented in Table \ref{FL_cmapps_results} and discussed henceforth. The top model performance (lowest \textit{RMSE} of 26.75) is achieved by the ‘Central’ model, which gives the benchmark RUL estimation performance for the ideal scenario of centrally accessing the entire train engine dataset for model training without privacy limitations. As the performance of the ‘Central’ model is comparable to the \textit{RMSE} values reported in existing literature \cite{chen2020machine}\cite{zheng2017long}, we can be confident that it is a high-quality benchmark for assessing the relative performance of the baseline client models and federated models.

\begin{figure}[!t]
\vspace{-1.2 em}
\centering
\subfloat[Heterogeneous clients]{\includegraphics[width=0.4\linewidth]{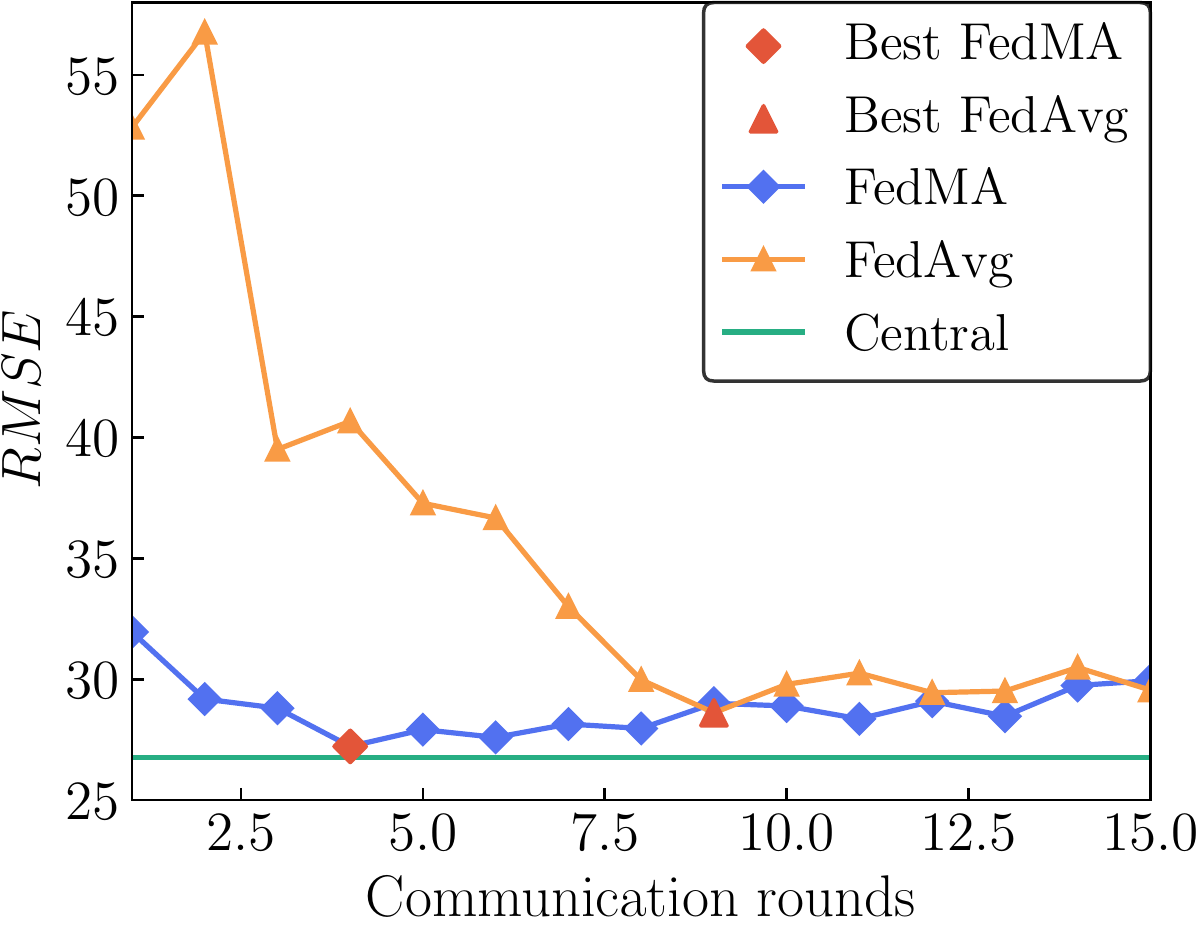}%
\label{het_clients}}
\hfil
\subfloat[Homogeneous clients]{\includegraphics[width=0.4\linewidth]{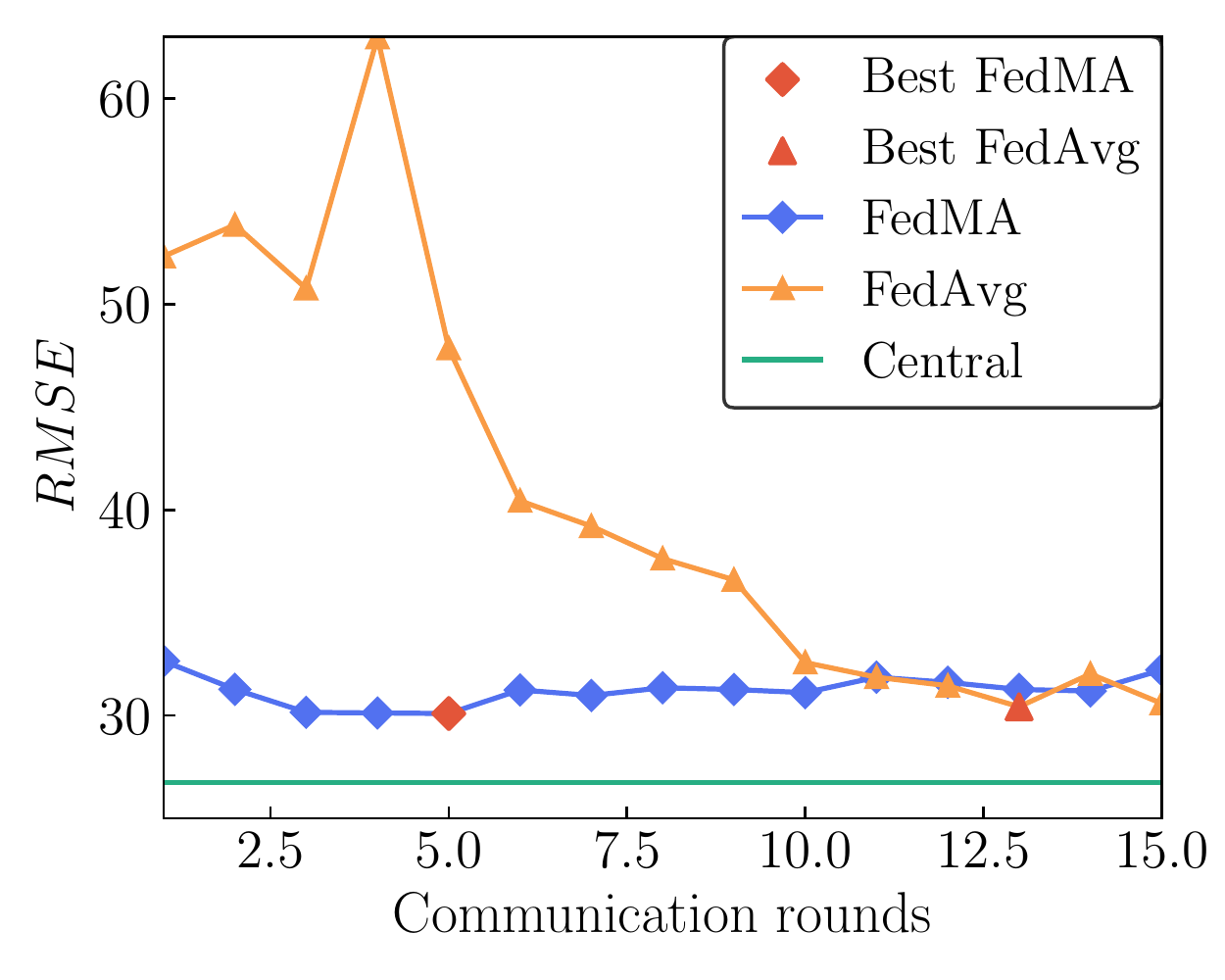}%
\label{homo_clients}}
\caption{Progression of the model accuracy for FedAvg and FedMA federated models under (a) heterogeneous and (b) homogeneous settings.}
\vspace{-1.6em}
\label{FL_results_cmapss}
\end{figure}

The performance of the baseline client models with no FL varies widely depending on the degree of client heterogeneity. While the model performance of homogeneous clients was fairly uniform (as evidenced by the narrow \textit{RMSE} range of 32.27 to 36.15), the heterogeneous clients had extremes, with very good models (HT2 with an \textit{RMSE} of 27.76) and very poor models (HT3 with an \textit{RMSE} of 44.87). Nonetheless, the federated models outperformed all the baseline client models, highlighting once again the crucial need for FL if high-accuracy machine health prognostics models are to be achieved under privacy constraints. Between the two FL methods, the FedMA federated model consistently topped FedAvg’s performance in both the homogeneous and heterogeneous settings, with its \textit{RMSE} in the heterogeneous setting being only 1.8\% lower  than the theoretical best performance established by the ‘Central’ model. According to \cite{wang2020federated}, FedMA performs better with higher quality local models. Thus, it is entirely consistent that FedMA’s outperformance to FedAvg was more pronounced in the heterogeneous setting, which had a high-quality local model (HT2). Due to its similarity matching algorithm, FedMA is able to benefit from the presence of high-quality local models to a greater extent as the feature extractors from good local models are not diluted by naive parameter averaging. Fig. \ref{FL_results_cmapss} plots the progression of the federated model accuracy for FedAvg and FedMA across the communication rounds. FedMA required less communication rounds for its best model, even in the homogeneous setting with only moderate quality local models, and converged to more optimal results than FedAvg.

\begin{figure}[!t]
\vspace{-2 em}
\centering
\subfloat[Baseline client models]
{\includegraphics[width=\linewidth]{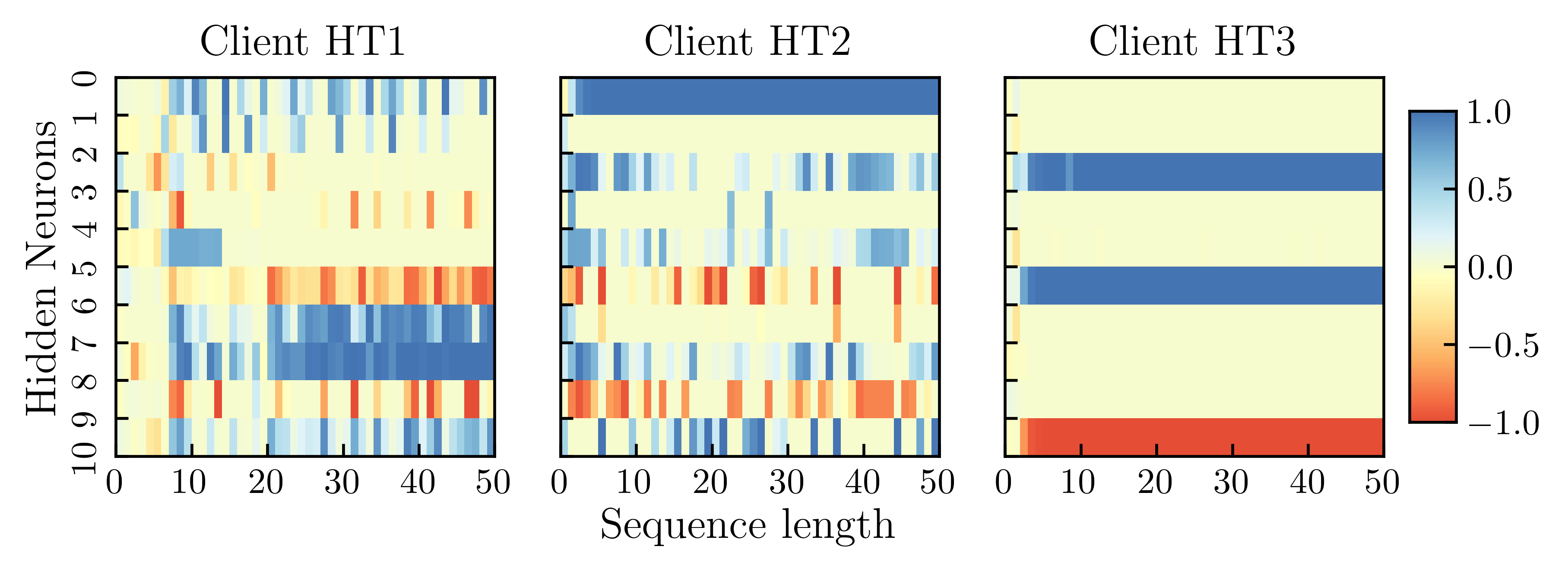}%
\label{before_perm}}
\hfil
\subfloat[Feature extractors of local models after first round of FedMA]{\includegraphics[width=\linewidth]{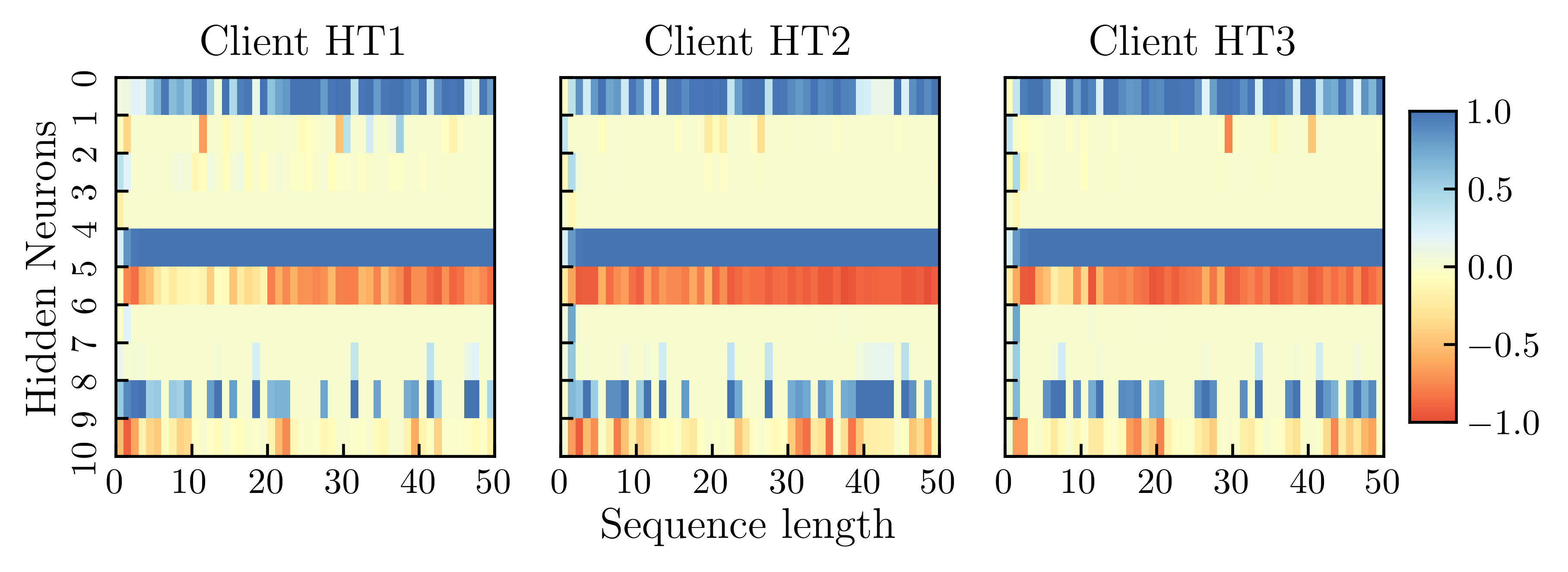}%
\label{after_perm}}
\caption{Visualization of feature extraction traits of local models in (a) the baseline setting without FL collaboration and after first round of (b) FedMA.}
\vspace{-1.6em}
\label{FL_interpretability}
\end{figure}

In Section III, FedMA’s neuron matching behavior was only described mathematically, and its \textbf{practical interpretability}  was not discussed. Thus, the feature extractors (neurons) of the local client models in the baseline scenario without FL, and after the first communication round of FedMA are visualized for the heterogeneous client setting in Fig. \ref{FL_interpretability}. Feature extractors are the LSTM layer’s output before it is fed to the fully connected dense layer. Only the first ten neurons of the 256 hidden neurons are plotted to allow for easy visual interpretation of the patterns. Starting with Fig. \ref{FL_interpretability}\subref{before_perm}, which depicts the feature extractors of the local models in baseline setup without FL, it is clear that each client’s coordinate-wise neurons have learnt significantly different feature extraction traits from their heterogeneous training data. Hence, performing coordinate-wise model parameter averaging on dissimilar traits, as FedAvg does, will dampen the feature extraction capabilities of the federated model. Fig. \ref{FL_interpretability}\subref{after_perm} plots the feature extractors amidst local model training after the first communication round of FedMA, which has already matched and aligned similar neurons index-wise with just one communication round.

\subsection{Complexity Analysis}
In this section, the computational complexity of the feature matching-based health prognostic model is briefly discussed, together with  some practical strategies for speed-ups. The computational complexity of an algorithm can refer to both space and time complexity \cite{cormen2022introduction}.

In a FL setting, the clients are edge devices with limited computational resources \cite{nguyen2021federated} to perform offline model training and online prognostics, i.e., inferencing. Thus, space complexity is a greater concern for the client side than the central server. Space complexity, in terms of memory consumption, is determined by factors such as input data size, number of model parameters, and the type of optimization algorithms utilized \cite{goodfellow2016deep}.  For training the local LSTM-based client models, the complexity per weight and time step for updating model weights is  $\mathcal{O}\left(1\right)$ \cite{hochreiter1997long} as the LSTM algorithm is local in time and space, i.e., each LSTM cell's output depends only on the previous cell's output and the current input. Furthermore, to reduce space complexity during training and inferencing, a light model architecture was chosen for the client model and the federated model, as mentioned in Section IV-A(2). For instance, our SoH estimation model has 68K parameters, which is a fraction of the 4.2M parameters of MobileNet \cite{howard2017mobilenets}, a popular parameter-efficient neural network for edge device deployment. Assuming parameters are stored in 32-bit floating-point numbers, the memory consumption of the model is approximately 2.7 Megabytes.

On the other hand, time complexity becomes the more critical factor in the federated model training phase as computing power is typically not a major constraint in the central server. The crux of the computational effort for the similarity-matched federated model development lies in the minimization of the matching cost matrix $\mathbf{P}$ via the Hungarian algorithm. If no hidden neurons are matched among clients, the worst case complexity per layer is $\mathcal{O}\left(D_{in} \cdot(J L)^2\right)$ for constructing $\mathbf{P}$ and $\mathcal{O}\left((J L)^3\right)$ for executing the Hungarian algorithm \cite{wang2020federated}. If all hidden neurons are matched, the best case complexity per layer is $\mathcal{O}\left(D_{in} \cdot L^2+L^3\right)$ \cite{wang2020federated}. In practice, the complexity is likely to be better than the worst case \cite{yurochkin2019bayesian} as indiscriminate growth of the federated model size is limited via the mechanism discussed in Section III-B(1). For practical implementation strategies, it is worth briefly noting that running time speed-up of the Hungarian algorithm can be achieved by using GPUs \cite{date2016gpu} and improved linear sum assignment solvers. More detailed information is available in \cite{lapsolver} \cite{pylapsolver}.

Time complexity is also of interest during online prognostics by clients as inferences often need to be performed in near real-time. However, in comparison with offline federated model training, the time complexity of online prognostics is  lower as it only involves model inferencing using the well-trained federated model parameters and test data. Specifically, the inference process, which entails a forward pass of our one-layer LSTM-based federated model, has a complexity of approximately $\mathcal{O}\left(D_{in} \cdot L + L^2\right)$.

\section{Conclusion}
This paper argues that, in an increasingly privacy-concerned era, federated learning (FL) is critical for allowing local clients to learn from a diverse range of decentralized data and build high-quality industrial health prognostic models. Particularly, we proposed a feature matching-based FL model for SoH and RUL estimation of critical machinery, and demonstrated its aptness for building accurate federated models amidst heterogeneous degradation data obtained under complex practical conditions. The proposed approach outperformed both the baseline client models without FL and the standard Federated Averaging model due to its algorithm averaging only client neurons with similar feature extraction traits, which ensures that the helpful feature extractors of local client models are carried over undiluted to the resultant federated model. Amidst the nascency of FL studies for time series regression-based machine health modeling, our study is pivotal for spurring research into effective parameter aggregation methods for FL-based industrial health prognostics.  Though the prognostic tasks studied in this paper use long short-term memory networks, the study can be extended to other deep neural networks in the future because the matched averaging is performed modularly layer-by-layer. Thus, as part of future works, the structurally simpler GRU network can be investigated for developing lighter health prognostics models in edge devices. Additionally, the technical feasibility of using an ensemble of privacy-preserving techniques such as differential privacy-enabled FL will also be studied for industrial health prognostics. Finally, as the current neuron matching algorithm allows for the nonparametric growth of the federated model and thus, susceptible to poisonous attacks, adversarial techniques to enhance the robustness and fault tolerance of the proposed model will be explored in future works.


%

%


\ifCLASSOPTIONcaptionsoff
  \newpage
\fi



\bibliography{mybib_fedma}
\bibliographystyle{IEEEtran}

\end{document}